\documentclass[letterpaper, 10 pt, conference]{ieeeconf}
\IEEEoverridecommandlockouts

\usepackage{cite}
\usepackage{amsmath,amssymb,amsfonts}
\usepackage{algorithm}
\usepackage{algorithmic}
\usepackage{graphicx}
\usepackage{textcomp}
\usepackage{xcolor}
\definecolor{weblink}{RGB}{0,82,155}   
\usepackage{colortbl}
\usepackage[font={small}]{caption}
\usepackage{booktabs}
\usepackage{multirow}
\usepackage{xspace}
\usepackage{microtype}
\usepackage[hidelinks]{hyperref}
\usepackage[hang,flushmargin]{footmisc}

\newcommand{\ourmodel}{D\textup{i}WALI\xspace}
\newif\ifarxiv
\arxivfalse
\newcommand{\figref}[1]{Fig.~\ref{#1}}
\newcommand{\tabref}[1]{Tab.~\ref{#1}}

\newcommand{\res}[2]{#1\,{\scriptsize$\pm$#2}}
\newcommand{\bres}[2]{\textbf{#1}\,{\scriptsize$\pm$#2}}

\def\BibTeX{{\rm B\kern-.05em{\sc i\kern-.025em b}\kern-.08em
    T\kern-.1667em\lower.7ex\hbox{E}\kern-.125emX}}

\begin{document}

\title{\LARGE \bf
\ourmodel: Diffusion Policy Learning in World Models \\ via Latent Action Inference from Human and Robot Play
}



\author{Akshay L Chandra$^{1}$, Iman Nematollahi$^{1}$, Lingyue Zhao$^{4}$, Daria Kulikova$^{1,3}$, \\ 
Angela P. Schoellig$^{4}$, Fabien Despinoy$^{2}$, Abhinav Valada$^{1}$
\thanks{$^{1}$Department of Computer Science, University of Freiburg, Germany $^{2}$Toyota Motor Europe $^{3}$ZUSE School ELIZA $^{4}$Technical University of Munich, Germany}%
\thanks{This work was funded by Toyota Motor Europe. Daria Kulikova is supported by the Konrad Zuse School of Excellence in Learning and Intelligent Systems (ELIZA) through the DAAD programme Konrad Zuse Schools of Excellence in Artificial Intelligence, sponsored by the Federal Ministry of Education and Research.}}%

\maketitle
\thispagestyle{empty}
\pagestyle{empty}

\begin{abstract}
Learning robot manipulation is constrained by the availability of action-labeled robot data,
for both imitation and world models. Humans can demonstrate
skills in seconds, but their demonstrations lack the required robot actions. 
We introduce \ourmodel, a framework that learns and improves diffusion policies within a
world model trained on human and robot play. \ourmodel infers latent
actions from consecutive frames using an inverse dynamics model, enabling
world model training without action labels, and grounds them in robot
commands through a decoder trained separately on robot play. Skill policies
are trained over latent actions from action-free human demonstrations and
improved with reinforcement learning in imagination. On the CALVIN
benchmark, \ourmodel policies trained solely from human demonstrations substantially
outperform baselines trained with robot demonstrations. Incorporating human play into
the world model training also yields stronger policies after fine-tuning than using the same
amount of robot play alone. Experiments on a physical robot show that skills
learned primarily from human demonstrations are further improved in
imagination with a world model trained on human and robot play. Together, these results demonstrate that human experience can support both manipulation skill acquisition and subsequent policy improvement, reducing reliance on action-labeled robot data. Videos, dataset and code are available at \href{https://diwali.cs.uni-freiburg.de}{\textcolor{weblink}{\texttt{https://diwali.cs.uni-freiburg.de}}}.
\end{abstract}

\section{Introduction}

Robot manipulation data is fundamentally constrained by robot time.
Demonstrations are typically collected through teleoperation, with a single
operator controlling a single robot in real time. As a result, every hour of
recorded data requires roughly one hour of both human and robot
time~\cite{openx2024}, leaving even the largest robot
datasets small by the standards of vision and language. World models~\cite{hafner2020dreamer, hafner2025dreamerv3}
learned from play can stretch this limited data budget. A single play dataset
collected in a scene can support a simulator in which policies are trained
and refined through imagined
rollouts~\cite{nematollahi2025lumos, chandra2025diwa}. However, both the play
data and the demonstrations must still include robot actions, since the
learned dynamics are conditioned on those actions.

\begin{figure}[t]
    \centering
\includegraphics[width=1.0\columnwidth]{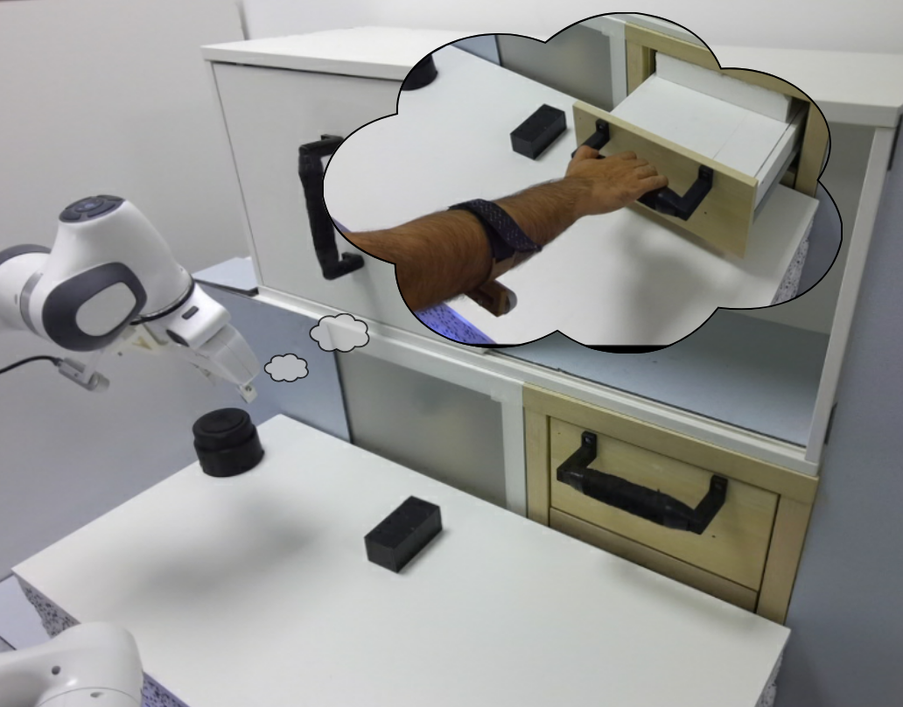}
    \caption{From human demonstrations to
robot skill: \textbf{\ourmodel} trains one world model on robot play and
action-free human play via latent actions inferred from consecutive frames. A diffusion
policy imitates human demonstrations in this latent space, improves in
imagination, and executes through a decoder trained on robot
play.\looseness=-1}
    \label{fig:cover}
    \vspace{-0.5cm}
\end{figure}

Human demonstrations are not subject to the same robot-time bottleneck. A
person can demonstrate a skill in seconds, yielding an order of magnitude
more demonstrations per hour than teleoperation. This
makes human interaction a far more scalable source of behavioral data, and
policies pre-trained on tens of thousands of hours of human data have
required only a few hours of robot data for
adaptation~\cite{zheng2026egoscale}.
However, these results rely on sensorized humans using tracked
hands~\cite{kareer2025egomimic},
gloves~\cite{zheng2026egoscale}, or handheld and wearable
grippers~\cite{honerkamp2025whole} to convert human motion into action
labels. Plain video of a person acting in the
robot's workspace provides no such action signal and, moreover, depicts a
different embodiment.

We seek a single model that learns from such video and subsequently serves
as a simulator in which a robot policy can be improved. Two lines of work
each address one half of this challenge. Latent action
models~\cite{ye2025latent, liang2025clam, bu2505learning} infer
transformations between consecutive frames of action-free video and use them
to pre-train policies, but the learned dynamics serve only as a training
signal rather than as a simulator. Action-conditioned world models provide
such a simulator~\cite{hafner2021dreamerv2, hafner2025training,
jiang2025world4rl, xiao2025worldenv}, but require action-labeled data. To our knowledge, no existing method combines both capabilities in a
single learned model that is used to improve a robot policy.

We close this gap with \textbf{\ourmodel}, which conditions the world model
on a \emph{latent action} inferred from two consecutive observations rather
than on a robot command (\figref{fig:cover}). An inverse dynamics model
infers this latent action, which the forward dynamics of a recurrent
state-space model use to predict the next latent state. Both are trained
end-to-end without action supervision, so the latent action captures whatever
aspects of a transition are needed to predict the next observation, with the
same representation used for human hands and robot grippers. A single model
can therefore learn scene dynamics from both human and robot play. A small
decoder, trained only on robot play and conditioned on the latent state,
grounds latent actions in executable robot commands, while a diffusion policy
is trained and fine-tuned entirely in the latent action space. The world
model thus serves both as the bridge between embodiments and as the simulator
in which the policy improves.

As imitation takes place in the latent action space, demonstrations no
longer require action labels. A skill shown by a human is encoded as latent
actions, imitated by the diffusion policy and decoded into
executable robot commands. On the CALVIN benchmark~\cite{mees2022calvin}, a
\ourmodel policy trained solely from human demonstrations outperforms
policies trained with a few robot demonstrations, whether based on a world
model conditioned on ground-truth robot actions or a latent action model
with a jointly trained decoder. Human play further broadens the world
model's training distribution, enabling imagination-based fine-tuning to
produce stronger policies than with the same amount of robot play alone.
Finally, on a real robot, policies learned primarily from
human demonstrations improve further through imagination, without
additional real-world interaction.

In summary, our contributions are:
\begin{enumerate}
    \item \textbf{A latent-action world model for human and robot play:} a
recurrent state-space model driven by an inverse dynamics model and trained
end-to-end without action supervision, together with an action decoder
trained separately on robot play and conditioned on the latent state.
The ablations show that both design choices are critical for making the latent
actions decodable.
\item \textbf{Skill policies from action-free human demonstrations:} diffusion
policies over latent actions that, when trained solely from human
demonstrations, outperform baselines trained with robot demonstrations.
\item \textbf{Human play for policy improvement in imagination:} action-free
human play is incorporated into the latent-action world model that serves
as the simulator, allowing human experience to shape the imagined rollouts
used for policy improvement rather than only policy pre-training. Policies
fine-tuned in this simulator outperform those fine-tuned with the same
robot play alone.
\item \textbf{Real-world validation:} the full pipeline, from
play data to a fine-tuned skill policy, deployed on a physical robot.\looseness=-1
\item We make the \textbf{code publicly available} at \url{https://diwali.cs.uni-freiburg.de}.
\end{enumerate}

\section{Related Work}
\label{sec:related}

{
\noindent\textit{Learning Robot Manipulation from Human Video}: 
Learning manipulation from human video requires a
representation that bridges the embodiment gap. Explicit
approaches build it from geometric cues extracted by
off-the-shelf perception: point tracks predicted from video and
followed by the robot~\cite{wen2023any, bharadhwaj2024track2act}, object
flow used as a cross-embodiment interface~\cite{xu2024flow,
heppert2026scaling}, and human hand poses retargeted to a robot
hand~\cite{zheng2026egoscale, luo2026beingh07}. These representations are
geometrically grounded, but inherit perception failures and impose a
hand-designed interface between embodiments.

Implicit approaches instead learn how human video should transfer to robot
behavior. They include visual encoders and skill embeddings pre-trained on
human video~\cite{karamcheti2023language, kim2025uniskill}, policies
co-trained on human and robot data~\cite{kareer2025egomimic,
punamiya2026egoverse}, video generators that predict how a task
unfolds~\cite{liang2025videopolicy}, and latent actions inferred between
frames. Latent actions can be learned from arbitrary video, but
must later be grounded in executable robot commands. \ourmodel belongs to this family, with one
key distinction: the latent actions learned from human video are also the
world model's action space, so the model bridging the embodiments
also serves as the simulator.}

{\parskip=2pt
\noindent\textit{Latent Action Models}: 
\label{sec:related_lam}
Vision-language-action models~\cite{kim2025openvla, black2024pi0, 
shukor2025smolvla, ding2024quarvla} scale with action-labeled robot
data~\cite{openx2024}. Latent action models offer a route beyond this
dependence. An inverse dynamics model infers a pseudo-action between
consecutive frames, and a forward model predicting the next frame from
it keeps the latent informative~\cite{schmidt2024lapo,
bruce2024genie, bu2026whatmatters, garrido2026latent}.
LAPA~\cite{ye2025latent}, UniVLA~\cite{bu2505learning},
Moto~\cite{chen2025moto}, and villa-X~\cite{chen2026villa} pre-train
policies on such latents from action-free video before fine-tuning on robot
actions. LAWM~\cite{tharwat2025latent} replaces the pairwise inverse model
with a forward instruction-conditioned policy and
LAOF~\cite{bu2026laof} introduces optical-flow constraints. Closest to our
setting, CLAM~\cite{liang2025clam} learns continuous latent actions with
inverse and forward dynamics and jointly trains an action decoder on a small
amount of labeled data. In these methods, however, the learned dynamics
mainly serve representation learning. \ourmodel instead learns
latent actions within a recurrent world model trained on human and robot
play, and trains the action decoder separately, conditioned on the latent
state; our ablations show both choices are critical for decodability.\looseness=-1}

{\parskip=2pt
\noindent\textit{Policy Improvement in Imagination}: 
Reinforcement learning can refine skills acquired through behavior
cloning~\cite{wang2019exploringmodelbasedplanningpolicy,
nematollahi2022robot, ren2025diffusion}, but
fine-tuning through physical interaction is costly. The
Dreamer family~\cite{hafner2020dreamer, hafner2021dreamerv2,
hafner2025dreamerv3, hafner2025training} replaces the environment with
learned dynamics and learns behaviors from imagined rollouts. DITTO and LUMOS~\cite{demoss2023ditto,
nematollahi2025lumos} learn imitation policies inside world models trained
on robot data, and DiWA, World4RL, and World-Env~\cite{chandra2025diwa,
jiang2025world4rl, xiao2025worldenv} post-train robot policies with
reinforcement learning inside offline world models. D2R~\cite{tharwat2026dream}
extends this to continual learning, rehearsing earlier tasks in
imagination to prevent forgetting. All
condition their dynamics on robot actions, requiring every training
sequence to include them. DGAC~\cite{chen2026dgac} instead learns dynamics
from human video, but improves the policy through real robot failures rather
than imagined interaction. DreamDojo~\cite{gao2026dreamdojo} conditions a
world model trained on human video on latent actions, but uses it to
evaluate and plan rather than to improve a policy. \ourmodel instead conditions the world model on inferred latent actions, so
human play trains the dynamics that later serve as the simulator, and the
policy is pre-trained and fine-tuned entirely in latent action space.\looseness=-1}
\vspace{-0.5cm}

\section{Problem Formulation}
\label{sec:problem}
\begin{figure*}[t]
    \centering
    \includegraphics[width=1.00\textwidth]{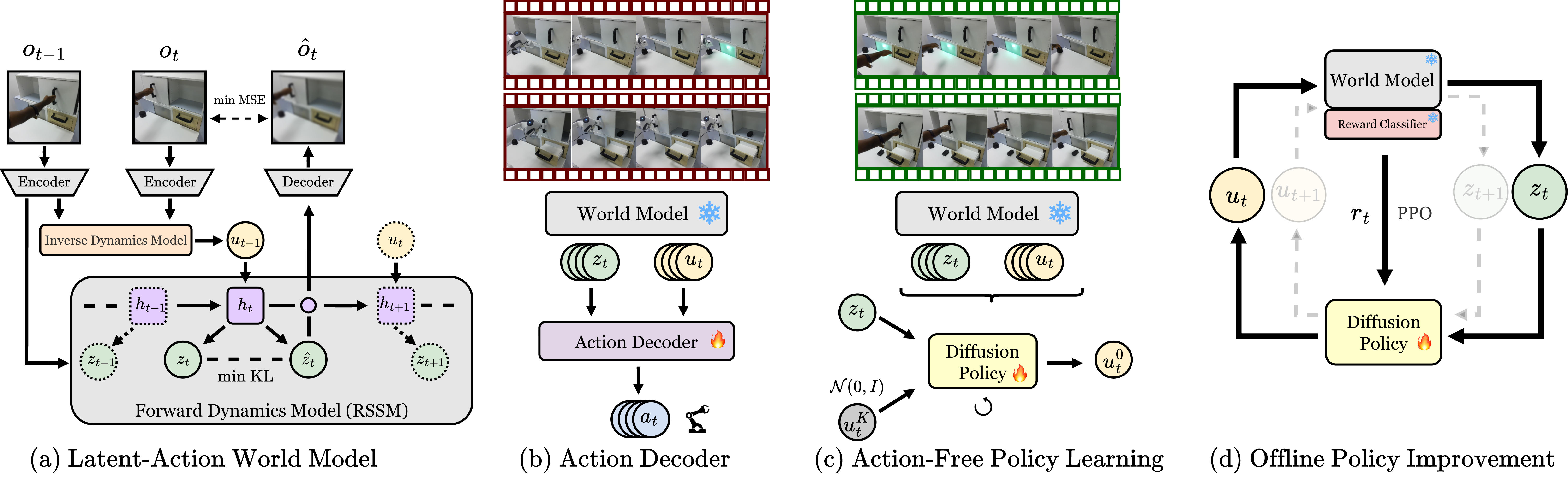}
    \vspace{-0.5cm}
    \caption{\textbf{\ourmodel} framework. (a) A latent-action world model is trained on robot and human play, where an inverse dynamics model infers a latent action $u_{t-1}$ from the embeddings of two consecutive observations, and the RSSM uses it in place of the robot command to advance its recurrent state, predict the prior $\hat{z}_t$, and reconstruct the observation. No action labels are used. (b) The frozen world model encodes robot play into latent states $z_t$ and latent actions $u_t$, from which an action decoder learns to predict the executed robot commands $a_t$. (c) Human demonstrations are encoded the same way, and a diffusion policy learns to denoise Gaussian noise into their latent actions conditioned on $z_t$. (d) The policy is fine-tuned with PPO on imagined rollouts, in which the frozen world model advances the latent state under the policy's latent actions, and a frozen reward classifier scores each state. Snowflakes mark frozen components, flames trainable ones.\looseness=-1}
    \label{fig:model_overview}
    \vspace{-0.3cm}
\end{figure*}
We study learning and improving robot manipulation skills when most of the
available interaction data comes from humans and therefore carries no robot
actions. We assume three offline datasets from the same workspace, each
observed by a fixed scene camera and an egocentric camera attached to the
acting gripper or hand: a task-agnostic robot play dataset
$\mathcal{D}^{R}_{\text{play}}$ of observation--action sequences, a human play
dataset $\mathcal{D}^{H}_{\text{play}}$ of observation sequences without
actions, and a small set of task-specific expert demonstrations
$\mathcal{D}_{\text{exp}}$ that are predominantly performed by a human and
hence action-free, with zero to a few by the robot. Observations $x_t$ are RGB
images only, so all datasets share one observation space. Our central
assumption is that human and robot act on the same scene, so the effect of a
motion is shared even when the command producing it is not.

We model the environment as a partially observable Markov decision
process $\mathcal{M}_{\text{env}} = (\mathcal{S}, \mathcal{A}, P, R, \gamma)$,
where $\mathcal{S}$ is the state-observation space, $\mathcal{A}$ the
continuous robot action space, $P(s_{t+1} \mid s_t, a_t)$ the dynamics,
$R(s_t, a_t)$ the reward, and $\gamma \in (0,1)$ the discount factor.
Rather than denoising robot actions, our diffusion policy
$\pi_\theta(u_t \mid s_t)$ generates a \emph{latent action}
$u_t \in \mathcal{U} \subset \mathbb{R}^{d}$ by sampling Gaussian noise
$\bar{u}_t^{K} \sim \mathcal{N}(0, I)$ and progressively denoising it:
\begin{equation}
\label{eq:diff_pi}
\bar{u}_t^{k-1} \sim \pi_\theta(\bar{u}_t^{k-1} \mid s_t, \bar{u}_t^{k}),
\quad k = K, \dots, 1,
\end{equation}
with $u_t = \bar{u}_t^{0}$. 
Since the robot executes actions in $\mathcal{A}$, a decoder $g_\xi$,
learnable from $\mathcal{D}^{R}_{\text{play}}$ alone, grounds latent actions
in robot actions, and the deployed policy is the composition
$g_\xi \circ \pi_\theta$. The diffusion policy is first pre-trained by
behavior cloning on $\mathcal{D}_{\text{exp}}$, and we then seek
\begin{equation}
\theta^\star = \arg\max_\theta\;
\mathbb{E}_{\tau \sim g_\xi \circ \pi_\theta}
\left[ \sum_{t=0}^{\infty} \gamma^t R(s_t, a_t) \right],
\end{equation}
where the expectation is over trajectories in $\mathcal{M}_{\text{env}}$.
Fine-tuning in $\mathcal{M}_{\text{env}}$ is impractical due to sample
complexity and safety. Instead, we learn a latent dynamics model
$P_\phi(z_{t+1} \mid z_t, u_t)$ on
$\mathcal{D}^{R}_{\text{play}} \cup \mathcal{D}^{H}_{\text{play}}$ and define
the world model MDP
\begin{equation}
\mathcal{M}_{\text{wm}} = (\mathcal{Z}, \mathcal{U}, P_\phi, R_\psi, \gamma),
\end{equation}
where $\mathcal{Z}$ is the learned latent space. Fine-tuning of $\pi_\theta$
is then performed entirely within $\mathcal{M}_{\text{wm}}$ through imagined
rollouts.

\section{\ourmodel Framework}
\label{sec:method}

We introduce \textbf{\ourmodel}, which trains in four stages
(\figref{fig:model_overview}): (1) a latent-action world model is learned from
robot and human play, (2) an action decoder is learned from robot play alone,
(3) a diffusion policy over latent actions is pre-trained on demonstrations,
and (4) the policy is fine-tuned within the world model and deployed
through the decoder.\ifarxiv{} Hyperparameters and pseudocode
(Algorithm~\ref{alg:diwali}) are given in the appendix.\fi

\subsection{Latent-Action World Model}
\label{sec:method_wm}
World models turn offline experience into a predictive model in which
behaviors can be learned without environment
interaction~\cite{hafner2020dreamer}. Following
DiWA~\cite{chandra2025diwa}, we use a recurrent state-space model
(RSSM)~\cite{hafner2021dreamerv2}. Separate convolutional encoders embed the
scene and egocentric images into $e_t$, a recurrent state $h_t$ summarizes
the past, a stochastic posterior $z_t$ is inferred from $e_t$ while a prior
$\hat{z}_t$ is predicted from $h_t$ alone, both categorical with
straight-through gradients~\cite{bengio2013estimating}, and a decoder
reconstructs $x_t$ from the model state $\hat{s}_t = (h_t, z_t)$. In action-conditioned
world models, the recurrent state is driven by the robot command $a_{t-1}$,
which human play does not carry. We therefore augment the RSSM with an inverse dynamics model
$I_\omega$, a tanh-squashed Gaussian conditioned on two consecutive
embeddings, whose sample replaces $a_{t-1}$:
\begin{equation}
\label{eq:model_components}
\begin{aligned}
    u_{t-1} &\sim I_\omega(u_{t-1} \mid e_{t-1}, e_t), \quad
    &u_{t-1} &\in \mathcal{U} = [-1, 1]^{d}, \\
    h_t &= f_\phi(\hat{s}_{t-1}, u_{t-1}), \quad
    &z_t &\sim q_\phi(z_t \mid h_t, e_t), \\
    \hat{z}_t &\sim p_\phi(\hat{z}_t \mid h_t), \quad
    &\hat{x}_t &\sim p_\phi(\hat{x}_t \mid \hat{s}_t),
\end{aligned}
\end{equation}
so that every sequence, human or robot, drives the dynamics. 
Encoders, inverse dynamics model,
dynamics, and decoder are trained end-to-end on
$\mathcal{D}^{R}_{\text{play}} \cup \mathcal{D}^{H}_{\text{play}}$ by
minimizing the negative ELBO:
\begin{align}
\label{eq:elbo}
\min_{\phi, \omega}\; \mathbb{E}_{q_\phi, I_\omega}\Big[
\sum_{t} & -\log p_\phi(x_t \mid \hat{s}_t) \\
& + \beta\, \mathrm{KL}\big(q_\phi(z_t \mid h_t, e_t)\,\|\,p_\phi(z_t \mid h_t)\big)
\Big]. \nonumber
\end{align}
Crucially, the objective contains no action term, so $I_\omega$ is trained
only by gradients that flow from the reconstruction and KL terms through
$f_\phi$ into its reparameterized samples. The latent action is therefore
whatever information about a transition the dynamics need to predict the next
observation, and its definition is identical for both embodiments by
construction.\looseness=-1

\subsection{Grounding Latent Actions in Robot Actions}
\label{sec:method_ad}
The world model operates in $\mathcal{U}$, but the robot executes commands
in $\mathcal{A}$, and nothing in Eq.~\eqref{eq:elbo} ties a latent action to
an executable command. The decoder $g_\xi$ supplies this link. It is the only
component that sees robot actions and is learned from
$\mathcal{D}^{R}_{\text{play}}$, the one dataset that pairs frames with
commands. 
Two design choices determine what $g_\xi$ learns. First, it is trained
after the world model, on
frozen features. If trained jointly, the robot-action loss would shape
$\mathcal{U}$ toward robot commands and pull or discard human transitions
that have no robot counterpart. Second, it is conditioned on the latent
state. A latent action encodes the effect of a motion, and the command that
produces a given effect depends on the robot's current configuration, which
only the state carries. Sec.~\ref{sec:ablations} ablates both choices. We
encode $\mathcal{D}^{R}_{\text{play}}$
with the frozen world model into $(z_t, u_t, a_t)$ triples, where
$a_t \in \mathbb{R}^{H \times d_a}$ is the chunk of $H$ robot commands
executed during one model step, and fit a deterministic MLP by least squares,
\begin{equation}
\label{eq:decoder}
\min_{\xi}\; \mathbb{E}_{\mathcal{D}^{R}_{\text{play}}}
\left[ \| g_\xi(u_t, z_t) - a_t \|_2^2 \right].
\end{equation}
$g_\xi$ stays frozen hereafter, even during fine-tuning.

\subsection{Policy Learning from Action-Free Demonstrations}
\label{sec:method_bc}
Skill policies are learned by imitating expert demonstrations, which
ordinarily must carry the robot's actions. This requirement makes them
expensive. With \ourmodel, the world model reads a latent action off any
two consecutive frames, so demonstrations can be imitated in $\mathcal{U}$
instead of $\mathcal{A}$, including human demonstrations, which are cheaper
to collect and make up most of $\mathcal{D}_{\text{exp}}$. Encoding $\mathcal{D}_{\text{exp}}$ with the frozen world model yields, for
every frame, a latent state $z_t$ and a latent action
$u_t \sim I_\omega(\cdot \mid e_t, e_{t+1})$. Robot demonstrations pass
through the same encoding, so their recorded actions are never read. The diffusion policy
$\pi_\theta(u_t \mid z_t)$ is trained to denoise noise into these latent
actions~\cite{chi2023diffusionpolicy}. Since $\mathcal{U}$ is bounded,
samples are clipped to it at no loss. Robot demonstrations therefore become optional, since a skill shown by a human is
executed by the robot via the inverse model, the policy, and the decoder.

\subsection{Policy Improvement in Imagination}
\label{sec:method_ft}
A cloned policy inherits the coverage and quality of its demonstrations.
Reinforcement learning inside the world model can improve it beyond them
without a single real interaction~\cite{chandra2025diwa}, but only as far as
the model's own coverage allows. The \ourmodel formulation makes that coverage
broad, because the model absorbs cheap human play, whereas a world model trained on robot play alone
can grow only through costly teleoperation. Improvement requires a task reward and a policy gradient through the
denoising process. 
The world model, trained on task-agnostic play, has no task reward. We label
the frames of each expert demonstration after task completion as successful,
train a binary classifier $C_\psi(z_t)$ on their latent states with these
frames as positives, and set $R_\psi(z_t, u_t) := C_\psi(z_{t+1})$.

For the policy gradient, we embed the $K$ denoising steps of the policy into
the world model MDP so that every step has a Gaussian
likelihood~\cite{ren2025diffusion, chandra2025diwa}: at world-model step $t$
and denoising step $k$,
\begin{equation}
\label{eq:ddmdp}
\begin{aligned}
\text{state: } &\bar{s}_{t,k} = (z_t, \bar{u}_t^{k}), \qquad
\text{action: } \bar{u}_t^{k-1}, \\
\text{reward: } &\bar{R}_{t,k} = \mathbb{1}[k = 1]\, R_\psi(z_t, \bar{u}_t^{0}),
\end{aligned}
\end{equation}
where the latent state is held fixed for $k > 1$ and advances by
$P_\phi(z_{t+1} \mid z_t, \bar{u}_t^{0})$ at $k = 1$. We optimize the policy
with PPO~\cite{schulman2017proximal} on imagined rollouts, using
denoising-step advantages
$\hat{A}(\bar{s}_{t,k}, \bar{u}_t^{k-1}) =
\gamma_{\text{denoise}}^{k}\,(\bar{r}_{t,k} - \hat{V}(z_t))$ and fine-tuning only the last $K_{\text{ft}}$ of the $K$ denoising steps~\cite{ren2025diffusion}.

Model exploitation~\cite{modelexploit} takes a specific form in a latent
action space. The dynamics and the decoder were trained only on samples from
$I_\omega$, so a latent action outside that distribution is an input neither
has seen, and a policy that finds them can score in
imagination and fail on the robot. We therefore regularize toward the pre-trained policy,
\begin{equation}
\label{eq:ft_loss}
\mathcal{L}_\theta = \mathcal{L}_{\text{PPO}}
- \alpha_{\text{BC}}\, \mathbb{E}_{\pi_{\theta_{\text{pre}}}}\left[\sum_{k=1}^{K_{\text{ft}}}
\log \pi_\theta(\bar{u}_t^{k-1} \mid z_t, \bar{u}_t^{k})\right],
\end{equation}
where the expectation is over latent states visited by the current policy
and denoising chains $\bar{u}_t$ drawn from the frozen
$\pi_{\theta_{\text{pre}}}$~\cite{torne2024reconciling}.
This keeps latent actions on the manifold the inverse model produces and the
decoder was trained on.

The BC term constrains each latent action, not the sequence, and
imagination rewards speed. With a sparse reward and a fixed rollout budget,
the fastest predicted success wins, pushing the policy toward abrupt
sequences the dynamics never saw. We therefore subtract from the terminal reward a penalty proportional to the
mean squared change between consecutive
latent actions over the episode. At deployment, the world model encodes the current observation into $z_t$,
$\pi_\theta$ samples a latent action, $g_\xi$ decodes it into a chunk of
robot commands, and the robot executes it.

\section{Experimental Evaluation}
\label{sec:experiments}

Our experiments address four questions. (i) Can a policy trained solely
from human demonstrations match or outperform policies trained with robot
demonstrations? (ii) Which design choices make latent actions decodable into
executable robot commands? (iii) Does human play in the world model yield better policies after
improvement in imagination than the same robot play alone? (iv) Does the full pipeline transfer to a physical robot? We
answer (i)--(iii) on the CALVIN benchmark~\cite{mees2022calvin} and (iv) in a
real-world setup.\looseness=-1

\begin{table*}[t]
\centering
\footnotesize
\setlength{\tabcolsep}{3.5pt}
\caption{\textbf{Skill learning from few or no robot demonstrations on CALVIN.} Success rates (\%, mean $\pm$ std over three seeds) of behavior-cloned policies on 11 tasks in five groups. \emph{Obs.} is the policy's input, raw RGB or the world model's latent state. \emph{Action} is what it denoises, robot end-effector deltas or latent actions. \emph{Demos} is the number of robot and human demonstrations per task. \ourmodel outperforms every baseline on average, and with no robot demonstrations at all it surpasses baselines given five (59.6\% vs.\ 35.2\% average success). Best result per column in bold.}
\label{tab:main}
\begin{tabular}{l|cc|cc|ccccc|c}
\toprule
\multirow{2}{*}{\textbf{Method}} & \multicolumn{2}{c|}{\textbf{Policy}} & \multicolumn{2}{c|}{\textbf{Demos}} & \multicolumn{6}{c}{\textbf{Success Rate (\%)}} \\
 & Obs. & Action & Robot & Human & \begin{tabular}[c]{@{}c@{}}\textbf{Drawer}\\{\scriptsize (2 tasks)}\end{tabular} & \begin{tabular}[c]{@{}c@{}}\textbf{Slider}\\{\scriptsize (2 tasks)}\end{tabular} & \begin{tabular}[c]{@{}c@{}}\textbf{Lightbulb}\\{\scriptsize (2 tasks)}\end{tabular} & \begin{tabular}[c]{@{}c@{}}\textbf{LED}\\{\scriptsize (2 tasks)}\end{tabular} & \begin{tabular}[c]{@{}c@{}}\textbf{Block}\\{\scriptsize (3 tasks)}\end{tabular} & \begin{tabular}[c]{@{}c@{}}\textbf{Avg.}\\{\scriptsize (11 tasks)}\end{tabular} \\
\midrule
\multirow{2}{*}{Diffusion Policy (ViT)} & \multirow{2}{*}{RGB} & \multirow{2}{*}{EE-$\Delta$} & 5 & 0 & \res{71.8}{5.8} & \res{38.8}{9.7} & \res{29.7}{9.7} & \res{28.0}{20.0} & \res{7.9}{12.2} & \res{35.2}{11.5} \\
 & & & 1 & 0 & \res{23.4}{23.3} & \res{7.9}{4.1} & \res{12.5}{10.9} & \res{7.7}{5.4} & \res{4.2}{8.4} & \res{11.1}{10.4} \\
\midrule
\multirow{2}{*}{DiWA} & \multirow{2}{*}{WM} & \multirow{2}{*}{EE-$\Delta$} & 5 & 0 & \res{39.0}{4.6} & \res{19.6}{9.8} & \res{32.6}{15.7} & \res{18.8}{10.7} & \res{7.8}{8.8} & \res{23.6}{9.9} \\
 & & & 1 & 0 & \res{1.1}{2.7} & \res{8.9}{6.4} & \res{2.6}{3.0} & \res{4.8}{7.4} & \res{3.5}{2.6} & \res{4.2}{4.4} \\
\midrule
\multirow{2}{*}{CLAM} & \multirow{2}{*}{WM} & \multirow{2}{*}{Latent} & 5 & 0 & \res{9.1}{7.7} & \res{7.8}{7.4} & \res{6.1}{6.9} & \res{3.8}{1.2} & \res{7.3}{6.2} & \res{6.8}{5.9} \\
 & & & 1 & 0 & \res{4.7}{5.0} & \res{3.3}{2.9} & \res{1.1}{1.7} & \res{2.6}{4.2} & \res{3.4}{5.2} & \res{3.0}{3.8} \\
\midrule
\multirow{3}{*}{CLAM-X} & \multirow{3}{*}{WM} & \multirow{3}{*}{Latent} & 5 & 50 & \res{15.9}{6.8} & \res{10.4}{3.2} & \res{8.4}{5.3} & \res{7.6}{3.8} & \res{5.4}{4.3} & \res{9.5}{4.7} \\
 & & & 1 & 50 & \res{12.9}{6.3} & \res{10.0}{3.4} & \res{6.1}{6.6} & \res{7.0}{3.5} & \res{5.0}{6.0} & \res{8.2}{5.2} \\
 & & & 0 & 50 & \res{13.1}{7.9} & \res{10.0}{5.0} & \res{7.1}{3.1} & \res{5.9}{3.6} & \res{4.8}{5.7} & \res{8.2}{5.1} \\
\midrule
\multirow{4}{*}{\textbf{\ourmodel} (Ours)} & \multirow{4}{*}{WM} & \multirow{4}{*}{Latent} & 5 & 0 & \bres{81.2}{3.7} & \res{48.5}{9.7} & \res{48.5}{9.6} & \res{52.0}{11.1} & \res{10.5}{8.5} & \res{48.1}{22.5} \\
 & & & 5 & 50 & \res{67.5}{17.9} & \res{74.8}{11.5} & \bres{81.9}{3.9} & \bres{62.4}{11.2} & \bres{21.5}{10.9} & \bres{61.6}{11.1} \\
 & & & 1 & 50 & \res{64.7}{18.9} & \bres{77.1}{14.1} & \res{80.1}{4.9} & \res{60.1}{7.6} & \res{17.2}{10.2} & \res{59.8}{11.2} \\
 & & & 0 & 50 & \res{69.3}{16.0} & \res{77.0}{11.9} & \res{77.4}{7.9} & \res{57.5}{9.2} & \res{16.8}{5.8} & \res{59.6}{10.1} \\
\bottomrule
\end{tabular}
\vspace{-0.3cm}
\end{table*}

\subsection{Simulation Setup}
\label{sec:exp_setup}
\begin{figure}[t]
    \centering
    \includegraphics[width=0.99\columnwidth]{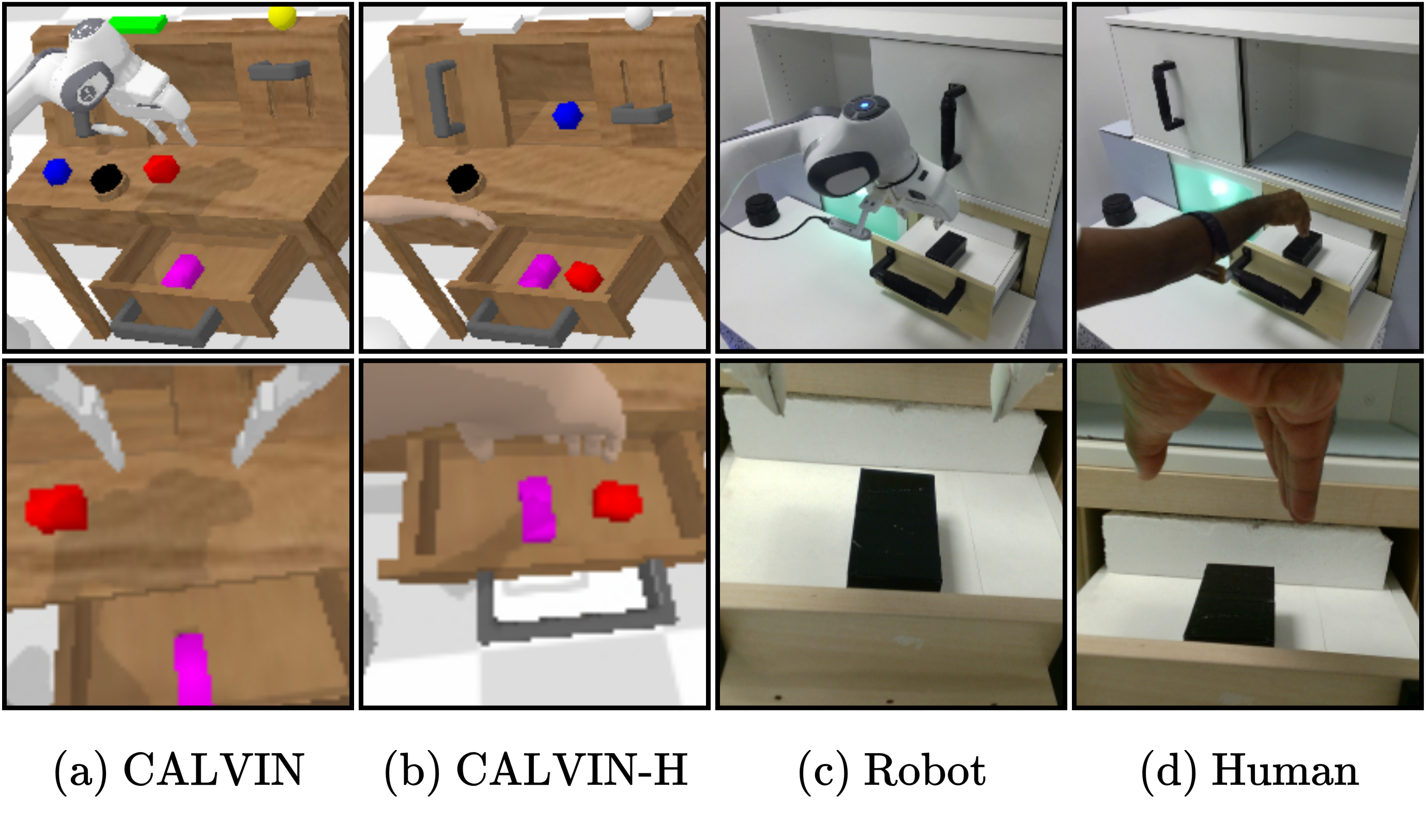}
    \vspace{-0.3cm}
    \caption{\textbf{Robot and human play.} Scene (top) and egocentric (bottom) views of simulated (a, b) and real-world (c, d) environments. (a) CALVIN and (b) CALVIN-H with a rendered human arm in place of the robot. (c) A real robot and (d) a real human acting in the real-world. In both domains, the human data has no action labels.}
    \label{fig:calvinh}
    \vspace{-0.3cm}
\end{figure}

We evaluate in environment \textit{D} of the CALVIN
benchmark~\cite{mees2022calvin}, where a 7-DoF Franka Emika Panda
manipulates a tabletop containing a drawer, a slider, a light bulb, an LED,
and colored blocks, observed by a fixed scene camera and a gripper camera.
CALVIN provides six hours of teleoperated play with action labels, which we
split evenly into two disjoint sets of episodes. The first serves as robot
play, $\mathcal{D}^{R}_{\text{play}}$. For the second, we replace the robot
with a rendered human arm and remove all action labels, yielding our human
play $\mathcal{D}^{H}_{\text{play}}$, which we call CALVIN-H (\figref{fig:calvinh}). Its egocentric view is
rendered from a camera mounted on the human wrist. The human arm differs from the robot
in appearance, occlusion patterns, and scene shading.

\textbf{Tasks and data.} We evaluate 11 tasks in five groups: opening and
closing the drawer, moving the slider left and right, switching the light
bulb and the LED on and off, and pushing, lifting, and rotating the pink
block. All world models are trained from $64\times64$ multi-view RGB frames for 900k
gradient steps on batches of 50 sequences of 50 model steps, each spanning
$H=5$ control steps. Latent actions have $d=16$ dimensions, and robot commands
$d_a=7$, an end-effector displacement and a gripper command. Each task
provides an expert set $\mathcal{D}_{\text{exp}}$ of 50 human demonstrations
and 0, 1, or 5 robot demonstrations. \ourmodel trains its world
model on $\mathcal{D}^{R}_{\text{play}} \cup \mathcal{D}^{H}_{\text{play}}$,
its action decoder on $\mathcal{D}^{R}_{\text{play}}$, and its policy on
$\mathcal{D}_{\text{exp}}$.\looseness=-1

\textbf{Baselines.} All methods train the same diffusion policy and differ
only in what it observes, what it predicts, and the data used to train its
representation. \emph{Diffusion Policy (ViT)} conditions the policy on multi-view RGB
observations encoded by a vision transformer and denoises robot
actions, without a world model. \emph{DiWA}~\cite{chandra2025diwa} instead conditions the policy
on the latent state of a world model trained on $\mathcal{D}^{R}_{\text{play}}$
with ground-truth actions, while likewise denoising robot actions. We
evaluate it both before and after fine-tuning in imagination. Since both
methods predict robot actions, they can imitate only demonstrations that
provide them and therefore train on robot demonstrations alone. \emph{CLAM}~\cite{liang2025clam} is re-implemented within our architecture
rather than directly replicated, allowing us to hold all components fixed
except the decoder training. Its policy denoises latent actions learned
through inverse and forward dynamics, while the action decoder is trained
jointly with the world model on $\mathcal{D}^{R}_{\text{play}}$.
\emph{CLAM-X} uses the same architecture but trains the world model on
$\mathcal{D}^{R}_{\text{play}} \cup \mathcal{D}^{H}_{\text{play}}$, with the
decoder loss applied to the robot half only. Neither variant conditions its
decoder on the latent state. Like \ourmodel, both train their policy on
$\mathcal{D}_{\text{exp}}$ and can imitate human as well as robot
demonstrations, all encoded by their frozen world model.\looseness=-1

\textbf{Protocol.} Each policy is evaluated from a fixed set of
task-specific initial configurations, taken from the first frames of CALVIN's
validation episodes, with one rollout per configuration. Success is
determined by CALVIN's task-completion criterion, and we report the mean and
standard deviation over three seeds. Diffusion policies use $K=20$ denoising
steps. Fine-tuning runs PPO for 670 iterations with 50 parallel imagined rollouts,
about one million world-model steps or five million control steps, and
updates only the final $K_{\text{ft}}=10$ denoising steps, using
$\gamma_{\text{denoise}}=0.99$ and a smoothness coefficient of 0.05. The
regularization weight $\alpha_{\text{BC}}$ in Eq.~\eqref{eq:ft_loss} is 0.1
for the open-drawer and move-slider-left tasks and 0.025 for all other
tasks, for DiWA and \ourmodel alike. For each
fine-tuned policy, the reward classifier is trained on the same
demonstrations used for policy pre-training, namely human and robot
demonstrations for latent-action policies and robot demonstrations alone for
DiWA.\looseness=-1

\subsection{Learning Skills from Human Demonstrations}
\label{sec:exp_skills}
\tabref{tab:main} evaluates whether a policy trained solely from human
demonstrations can match or outperform policies trained with robot
demonstrations. Without seeing a single robot demonstration of any task,
\ourmodel achieves 59.6\% average success. In comparison, the strongest
baseline given five robot demonstrations per task, Diffusion Policy (ViT),
reaches 35.2\%, while DiWA reaches 23.6\%. Action-free human demonstrations
are therefore sufficient to learn the skills on their own. Adding robot
demonstrations provides only marginal gains for \ourmodel, increasing success
from 59.6\% to 59.8\% with one demonstration per task and to 61.6\% with
five. The baselines, by contrast, degrade sharply when reduced from five
robot demonstrations to one, with Diffusion Policy (ViT) falling from 35.2\%
to 11.1\% and DiWA from 23.6\% to 4.2\%. These results show that \ourmodel's skills are driven primarily by the
human demonstrations, with robot demonstrations providing only a small
additional benefit.

Across task groups, \ourmodel achieves the best performance in every group. When trained with human demonstrations, it exceeds 57\% success on the drawer, slider, light bulb, and LED across all settings. Even with only five robot demonstrations and no human demonstrations, \ourmodel reaches 48.1\% average success, outperforming every baseline trained on the same demonstrations. This shows that the latent-action formulation provides a substantial benefit on its own, while human demonstrations contribute the additional gain. The block tasks remain the most challenging for all methods, as they require manipulating a small object from $64\times64$ observations. Even there, \ourmodel achieves the best result at 21.5\%.\looseness=-1

The two latent-action baselines perform substantially worse, remaining around 7--10\% across settings. Most
notably, CLAM-X uses exactly the same play data and demonstrations as
\ourmodel and shares the same policy architecture, yet achieves only 8.2\%
success compared with 59.6\% for \ourmodel when no robot demonstrations are
provided. The difference therefore lies in how the latent actions are
grounded into robot commands, which Sec.~\ref{sec:ablations} analyzes
through a sequence of targeted ablations.\looseness=-1

\subsection{Ablation Study}
\label{sec:ablations}
\begin{table}[t]
\centering
\footnotesize
\setlength{\tabcolsep}{6pt}
\caption{\textbf{Ablations.} Success (\%, mean $\pm$ std, three seeds) on 11 CALVIN tasks with 5 robot and 50 human demonstrations unless noted. (a) Decoder choices, added cumulatively to CLAM. (b) Auxiliary flow supervision. (c) Robot/human play mix at six hours.\looseness=-1}
\label{tab:ablations}
\begin{tabular}{lc}
\toprule
\multicolumn{2}{l}{\textbf{(a) Decoding latent actions}} \\
\midrule
CLAM: joint decoder, robot play                          & \res{6.8}{5.9} \\
+ human play (CLAM-X)                                    & \res{9.5}{4.7} \\
+ decoder trained separately                             & \res{18.7}{9.8} \\
+ decoder conditioned on latent state (\ourmodel)        & \bres{61.6}{11.1} \\
\midrule
\multicolumn{2}{l}{\textbf{(b) Auxiliary supervision for latent actions}} \\
\midrule
None (\ourmodel)      & \res{61.6}{11.1} \\
+ optical flow        & \res{61.3}{11.3} \\
+ point flow          & \bres{63.5}{10.9} \\
+ both                & \res{59.1}{9.3} \\
\midrule
\multicolumn{2}{l}{\textbf{(c) Robot / human mix of the play data}} \\
\midrule
1\% / 99\%              & \res{7.4}{5.4} \\
10\% / 90\%             & \res{36.2}{14.4} \\
25\% / 75\%             & \res{58.0}{11.6} \\
50\% / 50\% (\ourmodel) & \bres{61.6}{11.1} \\
\bottomrule
\end{tabular}
\vspace{-0.3cm}
\end{table}

\tabref{tab:ablations} examines the design choices that make latent actions
decodable into robot commands, the effect of auxiliary supervision, and how the
balance between human and robot play in the world model's training data
affects performance.\looseness=-1

\textbf{Decoding latent actions.} Block (a) starts from CLAM, whose action
decoder is trained jointly with the world model on robot play, and adds the
design choices from Sec.~\ref{sec:method_ad} one at a time,
using five robot and 50 human demonstrations. Extending the
jointly trained model with human play yields only a modest improvement, from
6.8\% to 9.5\%. As the decoder loss also shapes the latent space, human
transitions without corresponding robot actions cannot contribute cleanly to
learning the robot-action mapping. Training the decoder separately on frozen
world-model features instead leaves the latent action determined by the
dynamics alone and nearly doubles success to 18.7\%. Conditioning the decoder
on the latent state has the largest effect, raising success to 61.6\%. A
latent action captures the effect of a transition, but the robot command
required to realize that effect depends on the current configuration of the
robot relative to the scene, which the latent state provides. Together, these two design choices
account for the gap between CLAM-X and \ourmodel in \tabref{tab:main}.
Because the ablation is cumulative, each improvement is measured on top of
the preceding configuration.

\textbf{Auxiliary supervision.} Latent action models often supplement
their training objective with motion supervision~\cite{bu2026laof}. Block
(b) evaluates two such objectives in \ourmodel. An optical-flow head
predicts dense scene-camera flow from the current embedding and latent
action. It is trained without labels by warping the next frame with the
predicted flow and minimizing photometric reconstruction
error~\cite{nematollahi2020hindsight}, together with a
smoothness penalty. A point-flow head instead predicts the displacement of a
$7\times7$ grid of points, supervised by tracks precomputed with
CoTracker3~\cite{karaev2024cotracker3} over the full play dataset. Both heads
receive detached embeddings, so their gradients affect only the latent
action.

Neither objective yields a meaningful improvement. Success reaches 61.3\% with optical flow, 63.5\% with point flow, and
59.1\% with both, compared with 61.6\% without auxiliary supervision, and
all differences lie within one standard deviation. The small gain from point flow additionally requires
running an off-the-shelf tracker over hours of video to generate
supervision. The reconstruction and KL objectives therefore suffice to learn
useful latent actions, and \ourmodel uses no auxiliary supervision.

\textbf{Robot--human mix.} Block (c) varies the composition of the six-hour
play dataset from 1\% robot and 99\% human play to the default 50/50 mix.
Robot play provides the supervision needed to ground latent actions in
executable commands, since the action decoder is trained only on robot
data. With just 1\% robot play, corresponding to only a few minutes of
data, success falls to 7.4\%. Performance then rises sharply to 36.2\% at
10\% robot play and 58.0\% at 25\%, and saturates at the default mix with
61.6\%. Human play can therefore replace most, but not all, robot play.
Replacing three quarters of the robot play costs only 3.6 percentage
points, whereas replacing nine tenths costs 25.4 points. As changing the
mix affects both the learned dynamics and the action decoder, the sweep
does not isolate which component drives the gains.

\subsection{Policy Improvement in Imagination}
\label{sec:exp_finetune}
\begin{table}[t]
\centering
\footnotesize
\setlength{\tabcolsep}{2.2pt}
\caption{\textbf{Policy improvement in imagination.} Success rate (\%, mean $\pm$ std over three seeds) across 11 tasks before and after PPO fine-tuning in the world model. \emph{Play} denotes the world model's training data. \emph{Demos} are the robot (R) and human (H) demonstrations per task used to train the policy and reward classifier. Red marks higher-cost robot data, while green marks lower-cost or human data. \ourmodel outperforms DiWA trained on the same robot play with ten times as many robot demonstrations.}
\label{tab:finetune}
\begin{tabular}{llcccccc}
\toprule
\textbf{Method} & \textbf{Play} & \multicolumn{2}{c}{\textbf{Demos}} & \textbf{Before} & \textbf{After} & \textbf{Gain} \\
 & & R & H & & & \\
\midrule
DiWA & \cellcolor{red!12}100\% robot & \cellcolor{red!12}50 & 0 & \res{54.7}{9.7}  & \res{80.4}{4.9} & +25.7 \\
DiWA & \cellcolor{green!12}50\% robot  & \cellcolor{red!12}50 & 0 & \res{47.0}{10.9} & \res{69.5}{11.8} & +23.0 \\
\midrule
DiWA & \cellcolor{red!12}100\% robot & \cellcolor{green!12}5  & 0 & \res{28.2}{10.7} & \res{53.3}{11.7} & +25.1 \\
DiWA & \cellcolor{green!12}50\% robot  & \cellcolor{green!12}5  & 0 & \res{23.6}{9.9}  & \res{50.4}{12.2} & +26.8 \\
\textbf{\ourmodel} & \cellcolor{green!12}50\% robot, 50\% human & \cellcolor{green!12}5 & 0 & \res{48.1}{22.5} & \res{67.4}{13.7} & +19.3 \\
\textbf{\ourmodel} & \cellcolor{green!12}50\% robot, 50\% human & \cellcolor{green!12}5 & \cellcolor{green!12}50 & \bres{61.6}{11.1} & \bres{75.8}{5.9} & +14.2 \\
\bottomrule
\end{tabular}
\vspace{-0.3cm}
\end{table}

\tabref{tab:finetune} evaluates whether incorporating human play into the
world model yields better policies after improvement in imagination than
the same robot play alone. We denote DiWA configurations
by the amount of robot play used to train the world model and the number of
robot demonstrations as DiWA (play, R). All rows follow the fine-tuning
protocol of Sec.~\ref{sec:exp_setup}, differing only in the world-model
training data and policy demonstrations.

The closest comparison is between DiWA (50\%, 5) and \ourmodel with the
same five robot demonstrations and no human demonstrations. Both use the
same robot play, demonstrations, and PPO procedure. \ourmodel adds human
play to the world model and acts in latent action space. Its pre-trained
policy is already 24.5 points stronger, and although DiWA gains more from
fine-tuning, 26.8 against 19.3 points, \ourmodel ends 17 points higher at
67.4\%. Since this row uses no human demonstrations, the head start comes
from the world model itself, which is trained with human play and then
serves as the simulator for improvement. Adding 50 human demonstrations
raises \ourmodel to 75.8\%. Closing the gap to \ourmodel with robot data
alone is costly. Even with ten times as many robot demonstrations, DiWA
(50\%, 50) remains below \ourmodel, and only after doubling robot play does
DiWA (100\%, 50) surpass it by 4.6 points. With human demonstrations the
pattern repeats, and \ourmodel gains the least, 14.2 points, because its
pre-trained policy already outperforms every fine-tuned DiWA configuration
with five robot demonstrations. Each configuration trains its reward
classifier only on its own demonstrations, so none receives additional
reward supervision.\looseness=-1
\begin{figure}[t]
    \centering
    \includegraphics[height=3.4cm]{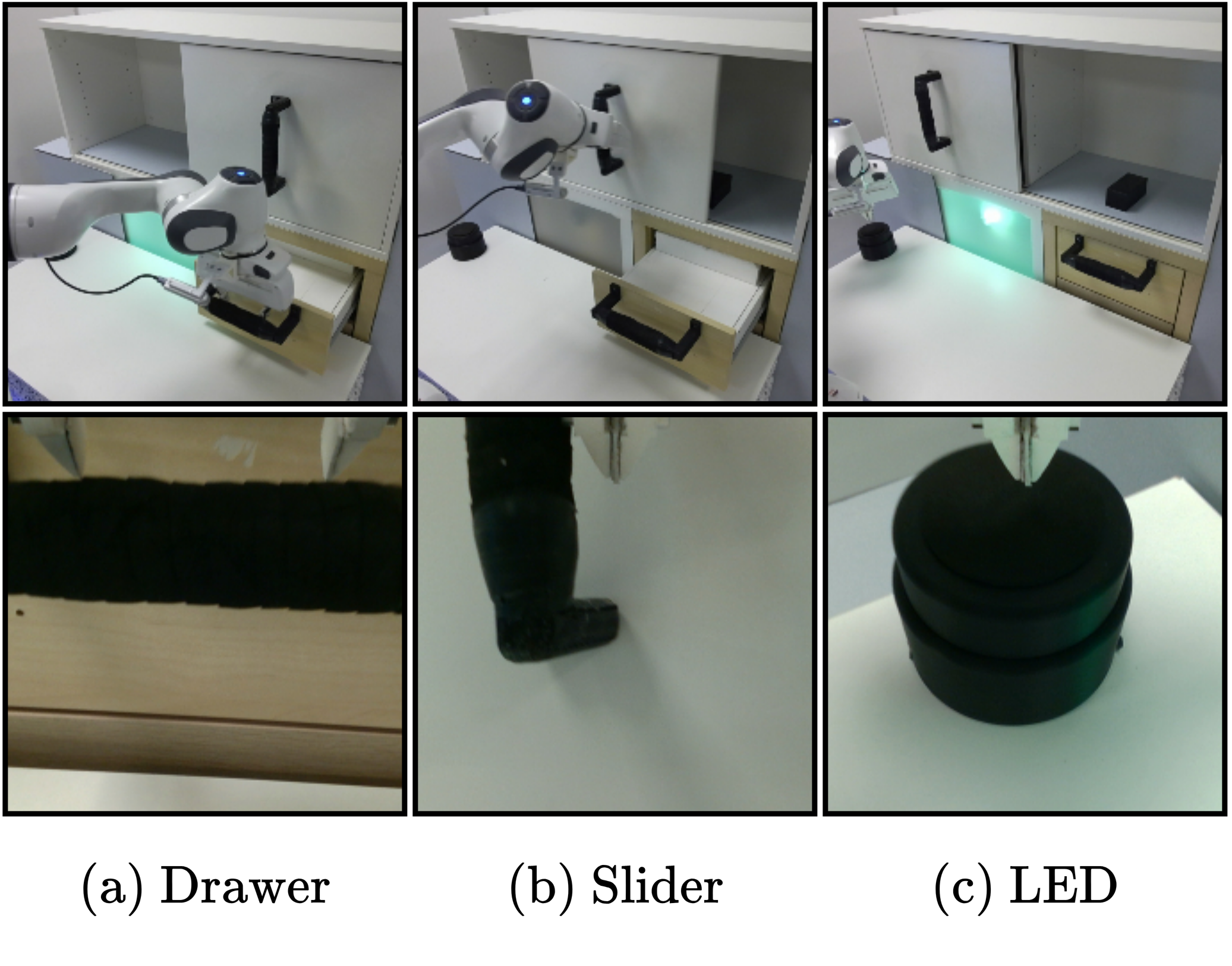}\hfill
    \includegraphics[height=3.4cm]{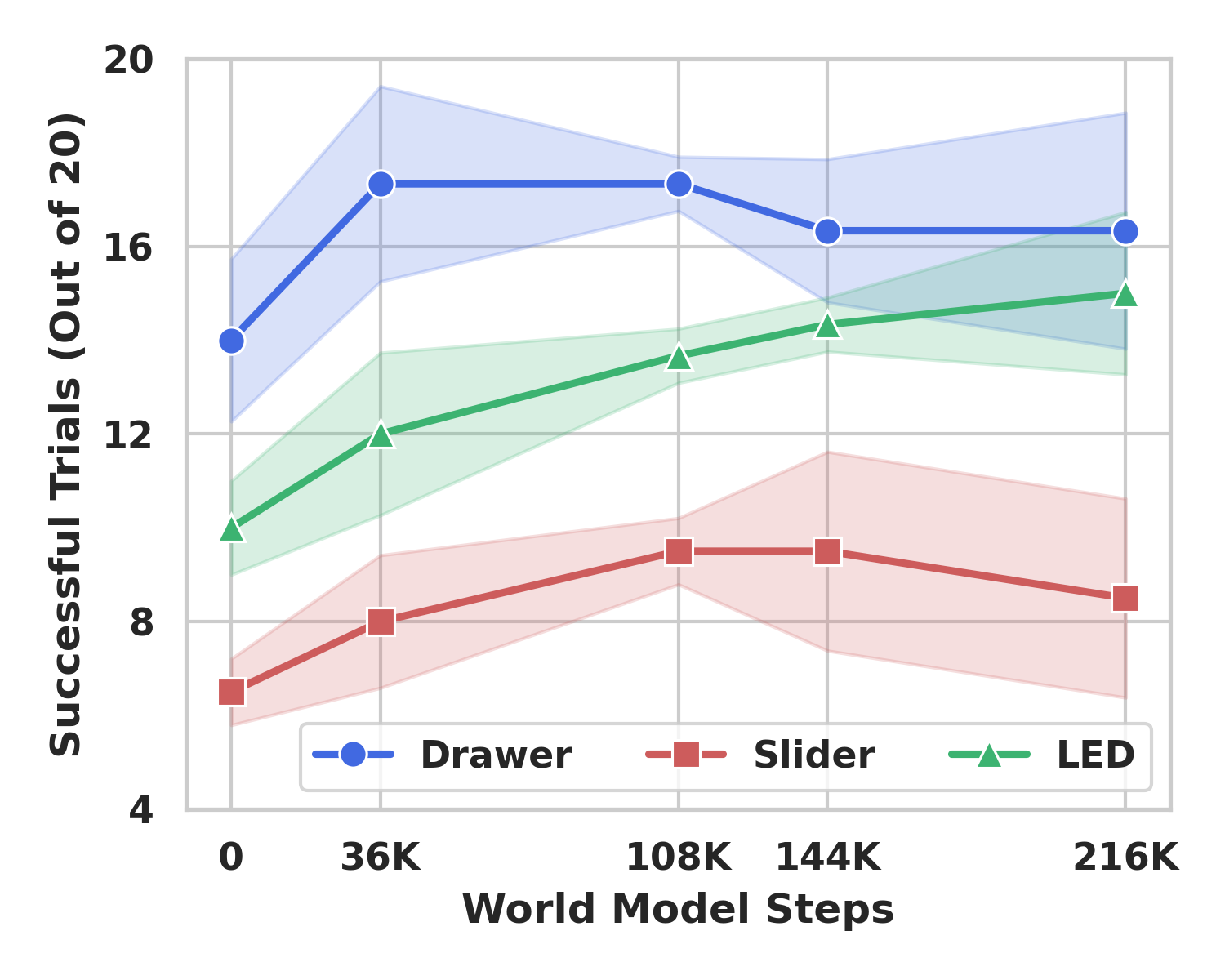}
    \caption{\textbf{Real-world results.} (a) The manipulation tasks used for evaluation. (b) Success before and after fine-tuning in imagination, averaged over 20 rollouts and three seeds. \ourmodel, trained on human and robot play, improves skills learned primarily from human demonstrations without any physical interaction.}
    \label{fig:realworld}
    \vspace{-0.3cm}
\end{figure}
\subsection{Real-World Results}
\label{sec:exp_real}
We evaluate \ourmodel on a Franka Emika Panda robot in a tabletop environment containing a cabinet with a drawer, a sliding door, and an LED controlled by a push button. The scene is observed by a static RGB camera and a gripper-mounted RGB camera (\figref{fig:cover}). Robot play comprises two hours of teleoperated interaction collected with a VR controller. Human play comprises two hours of a person manipulating the same objects in the same workspace, recorded by the static camera and a wrist-mounted camera corresponding to the gripper view, without action labels. Human motions therefore follow human rather than robot kinematics, and the wrist-mounted camera differs from the gripper camera in both placement and viewpoint. We evaluate three skills: opening the drawer, moving the slider to the left, and turning on the LED (\figref{fig:realworld}). For each skill, we collect 50 human demonstrations and five robot demonstrations to train the diffusion policy and reward classifier through the frozen world model, while the action decoder is trained on robot play alone. The world model is trained jointly on both play datasets and produces accurate long-horizon predictions on held-out trajectories. We then fine-tune the policies entirely in imagination for 216K world-model
steps, each spanning five control steps, or about one million control steps.

We evaluate each skill over 20 rollouts from fixed scene configurations and robot initial states, both before and after fine-tuning, and report results averaged over three seeds. Before fine-tuning, policies learned predominantly from human demonstrations already succeed in 53\% of rollouts. Imagination-based fine-tuning raises this to 70\%, improving performance on every skill. These results show that \ourmodel can improve real-world skills learned from human demonstrations using a world model trained in part on human play, without requiring additional physical interaction during policy improvement.

\section{Conclusion}
\label{sec:conclusion}
We presented \ourmodel, a world model driven by latent actions inferred
from consecutive frames, allowing human and robot play to train the same
dynamics. A decoder learned from robot play grounds these latent actions in
executable robot commands, while a diffusion policy is trained and improved
entirely in the latent action space. On CALVIN, policies trained solely from
human demonstrations outperform baselines trained with robot demonstrations.
Incorporating human play into the world model also leads to stronger
policies after imagination-based fine-tuning than using the same amount of
robot play alone. On a real robot, policies learned primarily from human
demonstrations further improve through imagination, without additional
real-world interaction.

Several future directions remain. \ourmodel currently learns from human play
collected in the robot's workspace. Extending it to in-the-wild human video
could expose the world model to broader experience. Latent actions also require robot play for grounding, motivating the question of how
little robot data is sufficient. Finally, policy improvement is limited by
a frozen world model. Updating the model with real rollouts from the
fine-tuned policy could further raise this ceiling.\looseness=-1








\bibliographystyle{IEEEtran}
\bibliography{main}

\newpage
\appendix
\setcounter{section}{0}
\setcounter{table}{0}
\setcounter{figure}{0}
\renewcommand{\thesection}{A.\arabic{section}}
\renewcommand{\thesectiondis}{\thesection.}
\renewcommand{\thesubsection}{A.\arabic{section}.\arabic{subsection}}
\renewcommand{\thesubsectiondis}{\thesubsection.}
\renewcommand{\thetable}{A.\arabic{table}}
\renewcommand{\thefigure}{A.\arabic{figure}}
\renewcommand{\thealgorithm}{A.\arabic{algorithm}}
\makeatletter
\def\section{\@startsection{section}{1}{\z@}{1.5ex plus 1.5ex minus 0.5ex}%
{0.7ex plus .5ex minus 0ex}{\normalfont\normalsize\itshape}}
\makeatother

\section{Hyperparameters and Training Details}
\label{sec:hyperparams}
Our model's architecture and training settings generally follow those of
DiWA~\cite{chandra2025diwa}, and we summarize them here alongside the differences.
All values are shared across simulation and real-world experiments unless otherwise specified.

\subsection{World Model}
\label{sec:hparams_wm}
The world model is a DreamerV2-style RSSM~\cite{hafner2021dreamerv2}, as in
DiWA~\cite{chandra2025diwa} and LUMOS~\cite{nematollahi2025lumos}. Separate
convolutional encoders embed the $64\times64$ scene and egocentric RGB images,
and their features are concatenated and fused by a fully connected layer into
the embedding $e_t$. The stochastic state $z_t$ consists of 32 categorical
variables with 32 classes each, and together with the 1024-dimensional
deterministic state $h_t$ it forms a 2048-dimensional model state. We use KL
balancing~\cite{hafner2021dreamerv2} and reset the recurrent state with a small
probability to cope with the scarcity of episode boundaries in play data.

\begin{figure}[t]
    \centering
    \includegraphics[width=0.853\linewidth]{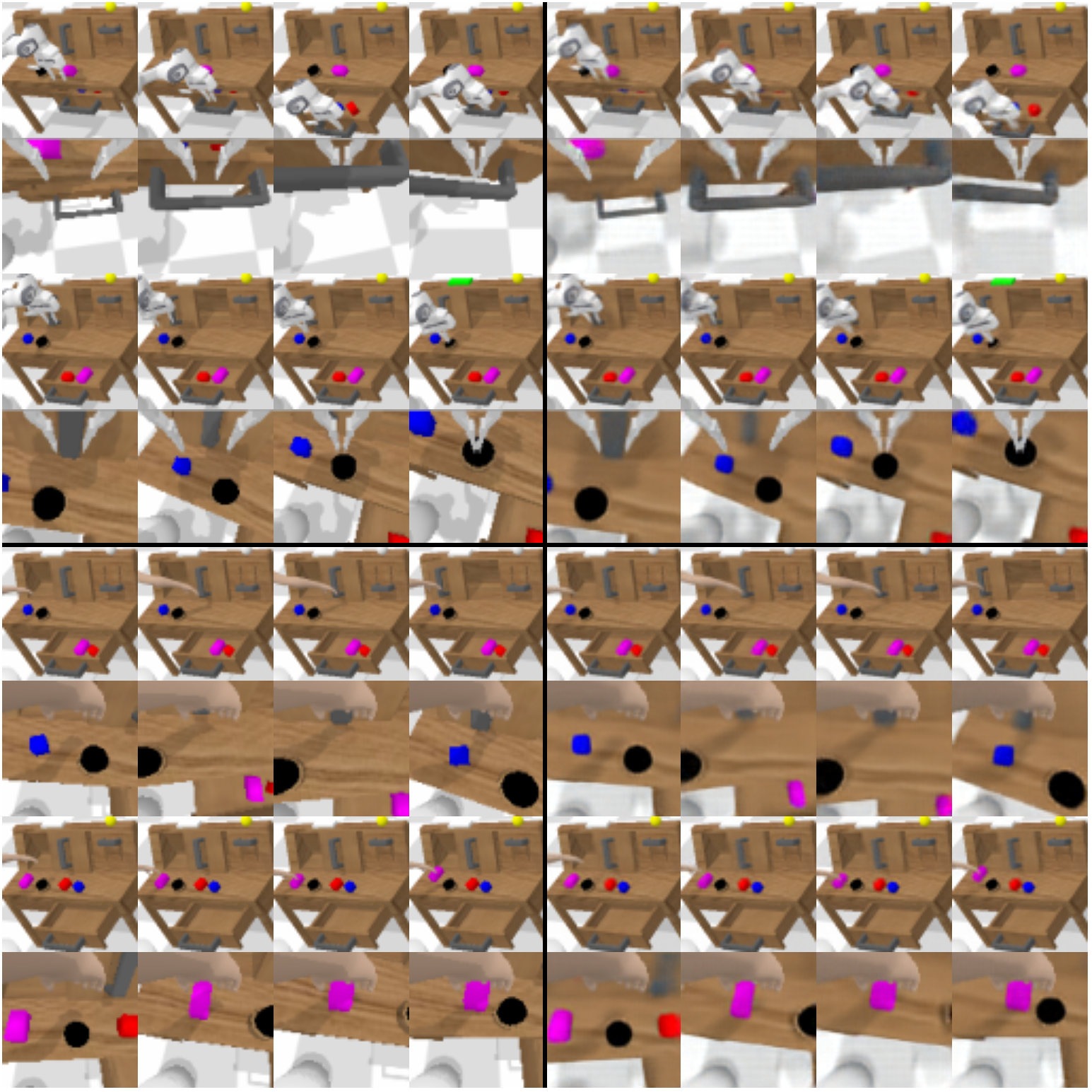}
    \caption{The world model rolls out futures from latent actions for the CALVIN robot (top) and the CALVIN-H human arm (bottom), matching ground-truth simulator rollouts (left) with imagined rollouts (right).}
    \label{fig:app_wm_calvin}
\end{figure}

\begin{figure}[t]
    \centering
    \includegraphics[width=0.853\linewidth]{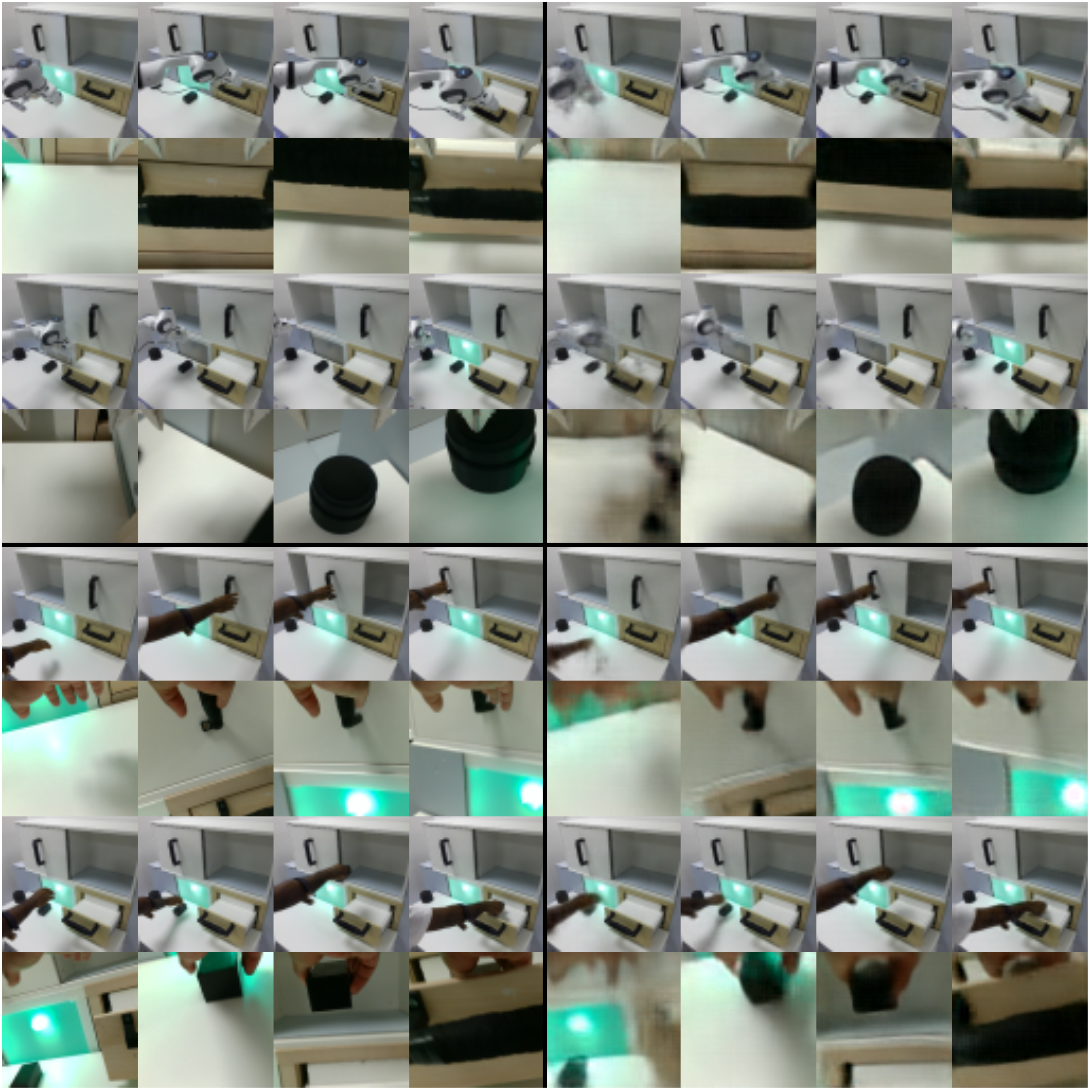}
    \caption{The world model rolls out futures from latent actions for the real robot (top) and the real human (bottom), matching ground-truth rollouts (left) with imagined rollouts (right).}
    \label{fig:app_wm_real}
\end{figure}

The inverse dynamics model $I_\omega$ replaces the robot command as the input to the recurrent dynamics model. It is an
MLP over the concatenated embeddings $(e_{t-1}, e_t)$ that outputs the mean and
log standard deviation of a Gaussian, whose reparameterized sample is squashed
by $\tanh$ into a $d=16$ dimensional latent action $u_{t-1} \in [-1,1]^{16}$.
$I_\omega$ is trained end-to-end with the rest of the model through
Eq.~\eqref{eq:elbo} and receives no action supervision. Each model step spans
$H=5$ control steps. Training sequences of 50 model steps are sampled from
random start indices within play episodes, and every batch mixes robot and
human play. All world models, including DiWA and CLAM baselines, are trained for
900k gradient steps. \tabref{tab:hparams_wm} lists the hyperparameters.
\figref{fig:app_wm_calvin} and \figref{fig:app_wm_real} show imagined rollouts
driven by inferred latent actions for both embodiments, in simulation and in
the real world, alongside the corresponding ground-truth rollouts.

\begin{table}[h]
\footnotesize
\centering
\caption{World model hyperparameters.}
\label{tab:hparams_wm}
\begin{tabular}{lcc}
\toprule
\textbf{Parameter} & \textbf{Symbol} & \textbf{Value} \\
\midrule
Image resolution & --- & $64\times64$ \\
Batch size & --- & 50 \\
Sequence length (model steps) & --- & 50 \\
Control steps per model step & $H$ & 5 \\
Deterministic state dimensions & --- & 1024 \\
Stochastic state (variables $\times$ classes) & --- & $32\times32$ \\
Latent action dimensions & $d$ & 16 \\
Inverse dynamics MLP & --- & $[256, 256, 256]$ \\
KL loss scale & $\beta$ & 1.0 \\
KL balancing coefficient & --- & 0.8 \\
RSSM reset probability & --- & 0.01 \\
Learning rate & --- & $3\times10^{-4}$ \\
Gradient clipping & --- & 100 \\
Adam epsilon & --- & $10^{-5}$ \\
Weight decay (decoupled) & --- & $5\times10^{-2}$ \\
Gradient steps & --- & 900k \\
\bottomrule
\end{tabular}
\end{table}

\subsection{Action Decoder}
\label{sec:hparams_ad}
The action decoder $g_\xi$ grounds latent actions in executable robot
commands. It takes the concatenation of the world model's latent state and the
latent action, both produced by the frozen world model, and outputs a chunk of
$H=5$ robot actions, each 7-dimensional (an end-effector displacement and a
gripper command), i.e., one chunk per model step. It is an MLP with layer
normalization and ELU activations, followed by a Gaussian head with separate
linear layers for the mean and the log standard deviation, the latter bounded
via $\tanh$ rescaling. The decoder is trained only on robot play
$\mathcal{D}^{R}_{\text{play}}$, the sole dataset that pairs observations with
actions, after the world model has been trained and frozen. It minimizes the
squared error between the predicted mean and the executed command chunk
(Eq.~\eqref{eq:decoder}), and at deployment only mean is executed. We optimize it with AdamW for 300k gradient steps, using a cosine-annealed
learning rate (\tabref{tab:hparams_ad}).

\begin{table}[h]
\footnotesize
\centering
\caption{Action decoder hyperparameters.}
\label{tab:hparams_ad}
\begin{tabular}{lc}
\toprule
\textbf{Parameter} & \textbf{Value} \\
\midrule
Input & latent state $\oplus$ latent action \\
Input dimensions & 2064 \\
Output dimensions & $5 \times 7$ \\
MLP dimensions & {[}1024, 512, 512, 256, 256{]} \\
Batch size & 50 \\
Optimizer & AdamW ($\beta_1{=}0.9$, $\beta_2{=}0.999$) \\
Learning rate (initial / minimum) & $3\times10^{-4}$ / $10^{-6}$ \\
Cosine annealing period & 100k steps \\
Gradient steps & 300k \\
Adam epsilon & $10^{-5}$ \\
Weight decay & $5\times10^{-2}$ \\
Training data & $\mathcal{D}^{R}_{\text{play}}$ only \\
\bottomrule
\end{tabular}
\end{table}

\subsection{Diffusion Policy}
\label{sec:hparams_dp}
The policy is a DDPM-based diffusion policy~\cite{chi2023diffusionpolicy} with $K=20$ denoising steps and
the same MLP architecture and training schedule as in
DiWA~\cite{chandra2025diwa} (\tabref{tab:hparams_dp}). The difference is its
output. Instead of robot end-effector deltas, it denoises the latent actions
$u_t$ that the frozen world model infers from consecutive demonstration frames,
conditioned on the 2048-dimensional latent state of the current observation.
Its samples are clipped to $\mathcal{U}=[-1,1]^{16}$ and decoded into a chunk
of $H=5$ robot commands by the frozen action decoder $g_\xi$. For each task, the
policy is trained on 50 human demonstrations and 0, 1, or 5 robot
demonstrations, all encoded by the frozen world model, so the recorded actions
of robot demonstrations are never used. The baselines in
\tabref{tab:main} train the same policy architecture with the same
hyperparameters and differ only in their inputs and outputs.

\begin{table}[h]
\footnotesize
\centering
\caption{Diffusion policy hyperparameters, shared across all tasks and methods.}
\label{tab:hparams_dp}
\begin{tabular}{lcc}
\toprule
\textbf{Parameter} & \textbf{Symbol} & \textbf{Value} \\
\midrule
Observation horizon & --- & 1 \\
Observation dimensions & --- & 2048 \\
Action dimensions (latent) & $d$ & 16 \\
Denoising steps & $K$ & 20 \\
MLP dimensions & --- & {[}512, 512, 512{]} \\
Training epochs & --- & 5000 \\
Batch size & --- & 256 \\
Initial / final learning rate (cosine) & --- & $10^{-4}$ / $10^{-5}$ \\
Weight decay & --- & $10^{-6}$ \\
EMA decay / start epoch / update freq. & --- & 0.995 / 20 / 10 \\
Demonstrations per task (human / robot) & --- & 50 / \{0, 1, 5\} \\
\bottomrule
\end{tabular}
\end{table}

\subsection{Latent Reward Classifier}
\label{sec:hparams_reward}
The reward classifier $C_\psi$ is identical to the latent reward estimator of
DiWA~\cite{chandra2025diwa}, including its architecture, contrastive and
cross-entropy objective, and hyperparameters, and we refer to that work for
details. The only difference is its training data. For each task, it is trained
on the same demonstrations as the policy, encoded by the frozen \ourmodel world
model, i.e., human and robot demonstrations for latent-action policies and
robot demonstrations alone for DiWA.

\subsection{Fine-tuning in Imagination}
\label{sec:hparams_ft}
Fine-tuning follows DiWA~\cite{chandra2025diwa}, which applies
DPPO~\cite{ren2025diffusion} to rollouts imagined by a frozen world model, and
uses the same PPO hyperparameters (\tabref{tab:hparams_ft}). Below, we summarize the
procedure and point out where \ourmodel departs from it. Algorithm~\ref{alg:diwali}
gives the full pseudocode for fine-tuning, and Algorithm~\ref{alg:diwali_deploy} for deployment.

\textbf{Imagined rollouts.} Each rollout starts from the model state
$\hat{s}_0 = (h_0, z_0)$ of an encoded initial observation. At every model step $t$,
the policy denoises Gaussian noise over $K$ steps into a latent action
$u_t = \bar{u}_t^{0}$, clipped to $\mathcal{U}$. The policy
outputs latent action $u_t$ which is fed
directly to the recurrent model, $h_{t+1} = f_\phi(\hat{s}_t, u_t)$, and the
next latent state is sampled from the prior $p_\phi(\hat{z}_{t+1} \mid h_{t+1})$.
Robot actions therefore never appear during fine-tuning, and the action decoder
$g_\xi$ is used only at deployment. Following Eq.~\eqref{eq:ddmdp}, only the last denoising step of
each model step is rewarded. Its reward is the classifier's score of the
predicted next state, $R_\psi(z_t, u_t) = C_\psi(z_{t+1})$. To avoid updates from
episodes rewarded by isolated classifier errors, an imagined episode counts as
successful, and terminates, only once $C_\psi$ flags success for three
consecutive model steps. Otherwise it ends after $T$ model steps.

\textbf{Smoothness penalty.} The penalty is computed directly on the latent
actions, without decoding them. At each model step, we measure the squared
change from the previous latent action,
\begin{equation}
\label{eq:app_jerk}
J_t = \mathbb{1}[t > 0]\, \|u_t - u_{t-1}\|_2^2,
\end{equation}
and when an episode of $T_e$ model steps terminates, we subtract
$c_{\text{smooth}} \cdot \frac{1}{T_e}\sum_{t=0}^{T_e-1} J_t$ from its final
reward, with $c_{\text{smooth}} = 0.05$ in all experiments. Although the
penalty never sees robot commands, it smooths the decoded behavior as well,
because the decoder maps nearby latent actions to nearby commands. Because the penalty
is applied once, after success has been determined, and is averaged over the
episode, it neither encourages shorter episodes nor masks the success signal.

\textbf{Policy and value updates.} As in DiWA, advantages are computed with
generalized advantage estimation (GAE) at the final denoising step of each model step,
\begin{equation}
\label{eq:app_gae}
\begin{aligned}
\hat{A}_{t,1} &= \textstyle\sum_{l \geq 0} (\gamma_{\text{WM}} \lambda)^l\, \delta_{t+l}, \\
\delta_t &= \bar{R}_{t,1} + \gamma_{\text{WM}} V_\nu(z_{t+1}) - V_\nu(z_t),
\end{aligned}
\end{equation}
and propagated to earlier denoising steps as
$\hat{A}_{t,k} = \gamma_{\text{denoise}}^{k} \hat{A}_{t,1}$. Only the last
$K_{\text{ft}}$ denoising steps are updated, while the first $K - K_{\text{ft}}$
are sampled from the frozen $\pi_{\theta_{\text{pre}}}$. The value function
$V_\nu(z_t)$ regresses the discounted return of the latent rewards. The policy
minimizes the behavior-regularized PPO loss of Eq.~\eqref{eq:ft_loss}. It differs
from DiWA in the direction of the regularizer. DiWA scores the current policy's
denoising chains under $\pi_{\theta_{\text{pre}}}$, whereas we maximize the
likelihood under $\pi_\theta$ of chains drawn from $\pi_{\theta_{\text{pre}}}$ at
the visited latent states, which explicitly keeps latent actions in the region
covered by $I_\omega$ and by the decoder's training data. We use
$\alpha_{\text{BC}} = 0.025$ for all tasks.

\textbf{Deployment.} At each step, the world model encodes the current
observation into $\hat{s}_t = (h_t, z_t)$, $\pi_\theta$ samples a latent action
$u_t$, and $g_\xi(u_t, z_t)$ decodes it into five robot commands, which are
executed before the next observation is encoded. The recurrent state is
advanced with the policy's latent action, $h_{t+1} = f_\phi(\hat{s}_t, u_t)$.

\begin{table}[h]
\footnotesize
\centering
\caption{Fine-tuning hyperparameters, shared across tasks and with DiWA unless noted.}
\label{tab:hparams_ft}
\begin{tabular}{lcc}
\toprule
\textbf{Parameter} & \textbf{Symbol} & \textbf{Value} \\
\midrule
Denoising steps & $K$ & 20 \\
Fine-tuned denoising steps & $K_{\text{ft}}$ & 10 \\
Control steps per latent action & $H$ & 5 \\
Parallel imagined rollouts & $N$ & 50 \\
World-model steps (CALVIN / real) & --- & ${\sim}$1M / 216k \\
Actor / critic learning rate & --- & $10^{-5}$ / $10^{-3}$ \\
Actor MLP & --- & {[}512, 512, 512{]} \\
Critic MLP & --- & {[}256, 256, 256{]} \\
World-model discount & $\gamma_{\text{WM}}$ & 0.999 \\
Denoising discount & $\gamma_{\text{denoise}}$ & 0.99 \\
GAE parameter & $\lambda$ & 0.95 \\
BC coefficient & $\alpha_{\text{BC}}$ & 0.1 / 0.025 \\
Smoothness coefficient & $c_{\text{smooth}}$ & 0.05 \\
Batch size & --- & 7500 \\
\bottomrule
\end{tabular}
\end{table}

\begin{algorithm}[t]
\caption{\ourmodel: Diffusion Policy Adaptation with World Models and Latent Actions}
\label{alg:diwali}
\footnotesize
\begin{algorithmic}[1]
\STATE Train world model $M_\phi$ with inverse dynamics model $I_\omega$ on play data $\mathcal{D}^{R}_{\text{play}} \cup \mathcal{D}^{H}_{\text{play}}$ using the ELBO objective (Eq.~\eqref{eq:elbo}), then freeze $M_\phi$ and $I_\omega$.
\STATE Train action decoder $g_\xi$ on latent states and latent actions of robot play $\mathcal{D}^{R}_{\text{play}}$ (Eq.~\eqref{eq:decoder}), then freeze $g_\xi$.
\STATE Encode expert demonstrations into latent states $z_t$ and latent actions $u_t \sim I_\omega(\cdot \mid e_t, e_{t+1})$ using the frozen world model.
\STATE Pre-train diffusion policy $\pi_\theta$ on latent expert demonstrations via behavior cloning.
\STATE Train reward classifier $C_\psi$ on the latent states of the expert demonstrations.
\STATE Initialize value function $V_\nu$.
\FOR{iteration $= 1, 2, \dots$}
    \STATE Initialize imagined rollout buffer $\mathcal{D}_\text{itr}$.
    \STATE Set $\pi_{\theta_\text{old}} = \pi_\theta$.
    \FOR{imagination episode $= 1, 2, \dots, N$ in parallel}
        \STATE Sample initial observation $x_0$ and encode to model state $\hat{s}_0 = (h_0, z_0)$.
        \STATE Initialize state $\bar{s}_{0,K} = (z_0, \bar{u}_0^K)$ in $\mathcal{M}_{\text{DD}}$, with $\bar{u}_0^K \sim \mathcal{N}(0, I)$.
        \FOR{imagined step $t = 0, \dots, T-1$, denoising step $k = K, \dots, 1$}
            \STATE Sample intermediate latent action $\bar{u}_t^{k-1} \sim \pi_{\theta_\text{old}}(\cdot \mid z_t, \bar{u}_t^k)$
            \IF{$k = 1$}
                \STATE Run final latent action $u_t = \bar{u}_t^0$ in the world model $M_\phi$
                \STATE Update recurrent state: $h_{t+1} = f_\phi(\hat{s}_t, u_t)$
                \STATE Sample next latent state: $z_{t+1} \sim p_\phi(\hat{z}_{t+1} \mid h_{t+1})$
                \STATE Predict reward: $\bar{R}_{t,1} = R_\psi(z_t, u_t)$
                \STATE Sample new noisy latent action: $\bar{u}_{t+1}^{K} \sim \mathcal{N}(0, I)$
                \STATE Set next state: $\bar{s}_{t+1,K} = (z_{t+1}, \bar{u}_{t+1}^K)$
            \ELSE
                \STATE Set reward: $\bar{R}_{t,k} = 0$
                \STATE Set next state: $\bar{s}_{t,k-1} = (z_t, \bar{u}_t^{k-1})$
            \ENDIF
            \STATE Add $(k, \bar{s}_{t,k}, \bar{u}_t^{k-1}, \bar{R}_{t,k})$ to $\mathcal{D}_\text{itr}$.
        \ENDFOR
    \ENDFOR
    \STATE Compute advantage estimates $\hat{A}_{t,1}$ using GAE (Eq.~\eqref{eq:app_gae}).
    \FOR{update $= 1, \dots, \text{num\_updates}$}
        \FOR{minibatch $= 1, \dots, B$}
            \STATE Sample $(k, \bar{s}_{t,k}, \bar{u}_t^{k-1}, \bar{R}_{t,k})$ and $\hat{A}_{t,1}$ from $\mathcal{D}_\text{itr}$.
            \STATE Compute denoising-discounted advantage $\hat{A}_{t,k} = \gamma_\text{denoise}^k \hat{A}_{t,1}$.
            \STATE Update $\pi_\theta$ using the regularized PPO loss (Eq.~\eqref{eq:ft_loss}).
            \STATE Update $V_\nu$ by regressing the discounted latent returns.
        \ENDFOR
    \ENDFOR
\ENDFOR
\STATE \textbf{return} fine-tuned policy $\pi_\theta$.
\end{algorithmic}
\end{algorithm}

\begin{algorithm}[!b]
    \caption{\ourmodel: Deployment with Latent Action Decoding}
    \label{alg:diwali_deploy}
    \small
    \begin{algorithmic}[1]
    \REQUIRE Frozen world model $M_\phi$, fine-tuned policy $\pi_\theta$, frozen action decoder $g_\xi$.
    \STATE Encode initial observation $x_0$ into model state $\hat{s}_0 = (h_0, z_0)$.
    \FOR{$t = 0, 1, \dots$ until task success or time limit}
        \STATE Sample latent action $u_t \sim \pi_\theta(\cdot \mid z_t)$ by denoising over $K$ steps.
        \STATE Decode into robot commands: $(a_t^{1}, \dots, a_t^{H}) = g_\xi(u_t, z_t)$.
        \STATE Execute the $H$ robot commands in the environment and observe $x_{t+1}$.
        \STATE Update recurrent state: $h_{t+1} = f_\phi(\hat{s}_t, u_t)$.
        \STATE Infer posterior latent state: $z_{t+1} \sim q_\phi(z_{t+1} \mid h_{t+1}, e_{t+1})$; set $\hat{s}_{t+1} = (h_{t+1}, z_{t+1})$.
    \ENDFOR
    \end{algorithmic}
    \end{algorithm}

\section{World Models with Auxiliary Flow Supervision}
\label{sec:aux_flow}
For the ablation in \tabref{tab:ablations}(b), we train three world-model
variants that add motion supervision to the latent actions, using optical
flow, point flow, or both. Each variant attaches a small decoder head that
maps the latent action $u_{t-1}$, together with the detached embedding
$\mathrm{sg}(e_{t-1})$, to a motion prediction for the transition from
$x_{t-1}$ to $x_t$, i.e., over one model step of $H=5$ control steps. Because
the embedding is detached, the auxiliary gradients reach the encoder only
through the inverse dynamics model $I_\omega$. All other components and
hyperparameters match Sec.~\ref{sec:hparams_wm}, and the auxiliary terms are
added to the negative ELBO of Eq.~\eqref{eq:elbo}:
\begin{equation}
\label{eq:app_aux}
\mathcal{L} = \mathcal{L}_{\text{ELBO}}
+ \lambda_{\text{OF}} \mathcal{L}_{\text{OF}}
+ \lambda_{\text{smooth}} \mathcal{L}_{\text{smooth}}
+ \lambda_{\text{PF}} \mathcal{L}_{\text{PF}},
\end{equation}
where the optical-flow terms are disabled in the point-flow variant and vice
versa, and the combined variant uses all of them.

\begin{figure}[t]
    \centering
    \includegraphics[width=0.9\linewidth]{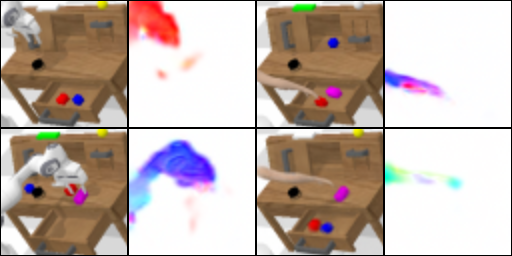}\\[2pt]
    {\footnotesize (a) Optical flow}\\[4pt]
    \includegraphics[width=0.9\linewidth]{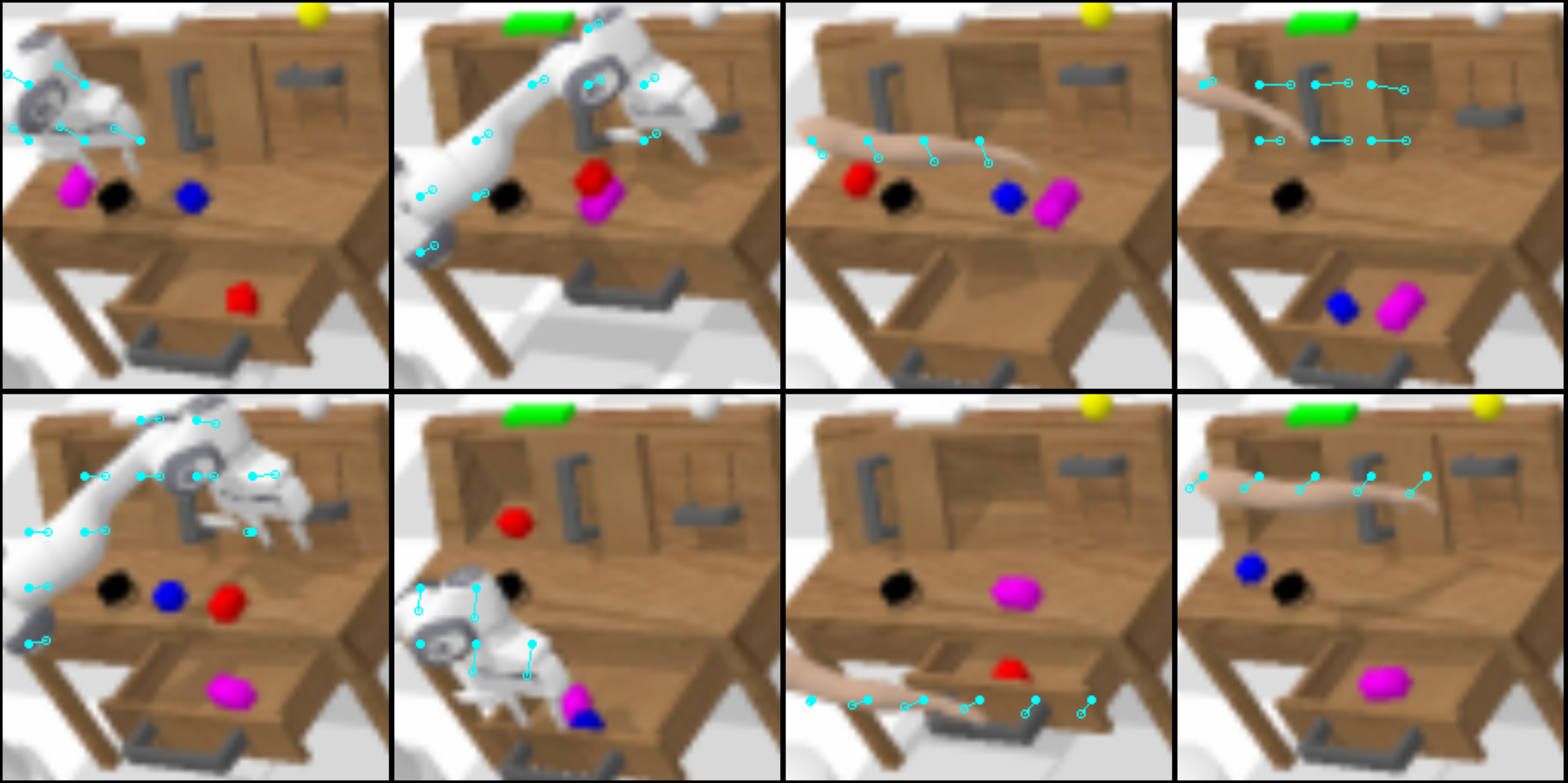}\\[2pt]
    {\footnotesize (b) Point flow}
    \caption{Auxiliary flow predictions decoded from latent actions on the
    static camera for robot play (left pairs) and human play (right pairs).
    Each pair shows, at different environment states, the observation and the flow prediction: (a) dense optical flow in
    standard color coding and (b) predicted point displacements overlaid on
    the observation.}
    \label{fig:app_flow_examples}
\end{figure}

\textbf{Optical flow.} The head predicts a dense flow field
$F_{t-1} \in \mathbb{R}^{64\times64\times3}$ and is trained without labels (\figref{fig:app_flow_examples}a).
The next frame is backward-warped with the predicted flow,
$\mathcal{W}(x_t, F_{t-1})$, using bilinear sampling, and compared with the
previous frame under an L1 photometric loss,
\begin{equation}
\label{eq:app_photo}
\mathcal{L}_{\text{OF}} = \sum_{p} \big\| \mathcal{W}(x_t, F_{t-1})(p) - \mathrm{sg}\big(x_{t-1}(p)\big) \big\|_1,
\end{equation}
summed over pixels $p$ and color channels. To regularize the flow in
textureless regions, we add an edge-aware second-order smoothness term
weighted by the image gradients of $x_{t-1}$,
\begin{equation}
\label{eq:app_smooth}
\mathcal{L}_{\text{smooth}} = \sum_{p} \sum_{d \in \{x, y\}}
\exp\!\big(-\big|\partial_d x_{t-1}(p)\big|\big)\,
\big|\partial_d^2 F_{t-1}(p)\big|.
\end{equation}
We use $\lambda_{\text{OF}} = 0.2$ and $\lambda_{\text{smooth}} = 0.01$.
When flow is also predicted for the gripper camera, each term is a weighted sum
over the two views, $\rho\,\mathcal{L}^{\text{static}} + (1-\rho)\,\mathcal{L}^{\text{gripper}}$,
with $\rho = 0.66$.

\textbf{Point flow.} We first build a point-tracking dataset for both robot
and human play. For every pair of consecutive steps, i.e., frames $H=5$
control steps apart, we track a $7\times7$ grid of equally spaced query points
with the official implementation of CoTracker3~\cite{karaev2024cotracker3} and
store their displacements $\Delta P_{t-1} \in \mathbb{R}^{49\times2}$ in
normalized image coordinates. The head predicts these displacements
$\widehat{\Delta P}_{t-1}$ in pixels (\figref{fig:app_flow_examples}b), and is supervised with a squared error
against the labels rescaled to the observation size $W = 64$,
\begin{equation}
\label{eq:app_pf}
\mathcal{L}_{\text{PF}} = \frac{1}{2} \sum_{i=1}^{49}
\big\| \widehat{\Delta P}_{t-1,i} - W\,\Delta P_{t-1,i} \big\|_2^2 .
\end{equation}
When the gripper camera is included, the static and gripper losses are
combined with separate weights. We use $\lambda_{\text{PF}} = 1.0$.

Both optical flow and point flow worked best when the auxiliary losses were
computed on the static camera only, leaving out the gripper view, and the
results in \tabref{tab:ablations}(b) use this configuration.
\figref{fig:app_flow_examples} shows example predictions of both heads at different environment states.

\section{Real-World Data Collection and Preprocessing}
\label{sec:real_data}
\begin{figure}[t]
    \centering
    \begin{minipage}[t]{0.571\linewidth}
        \centering
        \includegraphics[width=\linewidth]{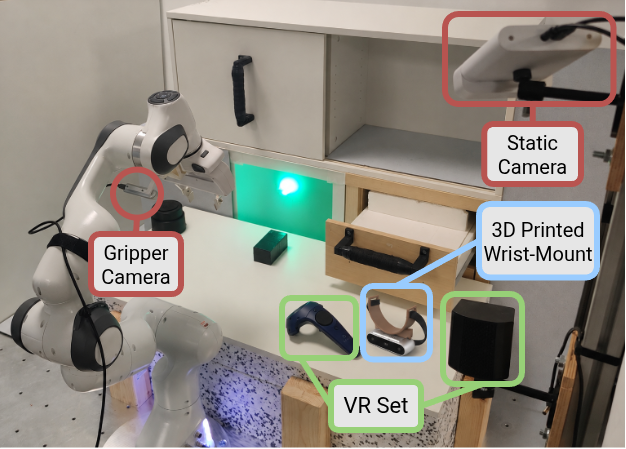}\\[2pt]
        {\footnotesize (a) Real-World Setup}
    \end{minipage}\hfill
    \begin{minipage}[t]{0.409\linewidth}
        \centering
        \includegraphics[width=\linewidth]{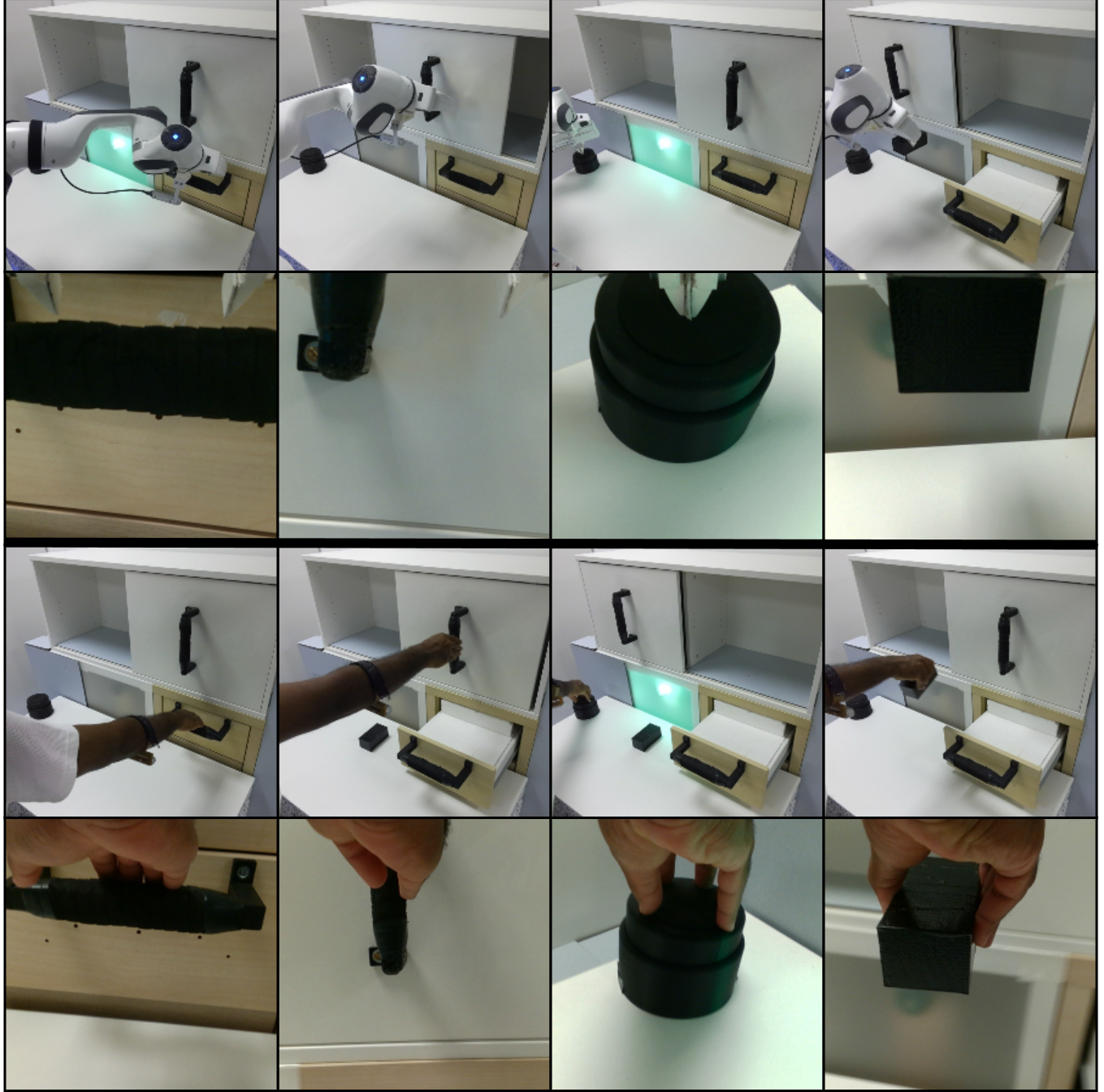}\\[2pt]
        {\footnotesize (b) Real-World Observations}
    \end{minipage}
    \caption{(a) Real-world setup showing the Franka Emika Panda, VR
    teleoperation interface, camera placements, the 3D-printed wrist mount used
    for human play (b) Example static and egocentric observations from
    robot play (top) and human play (bottom).}
    \label{fig:app_real_collection}
\end{figure}

The real-world setup consists of a Franka Emika
Panda robot teleoperated with an HTC VIVE Pro headset in a tabletop
environment containing a cabinet with a drawer, a sliding door, a push button
that switches an LED on, and a black cube (\figref{fig:app_real_collection}a). We
record RGB images of the full scene from a static Azure Kinect camera and close-up views
from a wrist-mounted Realsense D415 camera, both at $224\times224$
pixels, along with proprioceptive signals and the absolute control commands
sent to the robot. Relative actions are computed as differences between
consecutive absolute commands.

\textbf{Human play.} To collect human play, we remove the robot from the scene
and a person interacts with the same objects in the same workspace. The
egocentric view is recorded by a camera on a custom 3D-printed wrist mount
worn by the teleoperator, which places the camera at a position
roughly corresponding to the robot's gripper camera
(\figref{fig:app_real_collection}b). Human play carries no action labels.

\textbf{Preprocessing.} Both streams are recorded at 30\,Hz. To roughly match
the per-step motion magnitude across embodiments, we temporally downsample
robot data to 8\,Hz and human data to 15\,Hz, since human teleoperators
typically move faster than the robot. All images are resized to
$64\times64$ pixels and converted from integers in $[0, 255]$ to floats in
$[0, 1]$.

\end{document}